\PassOptionsToPackage{unicode}{hyperref}
\PassOptionsToPackage{hyphens}{url}
\PassOptionsToPackage{dvipsnames,svgnames*,x11names*}{xcolor}
\documentclass[
  american,
  letterpaper,
]{article}
\usepackage{amsmath,amssymb}
\usepackage{lmodern}
\usepackage{ifxetex,ifluatex}
\ifnum 0\ifxetex 1\fi\ifluatex 1\fi=0 % if pdftex
  \usepackage[T1]{fontenc}
  \usepackage[utf8]{inputenc}
  \usepackage{textcomp} % provide euro and other symbols
\else % if luatex or xetex
  \usepackage{unicode-math}
  \defaultfontfeatures{Scale=MatchLowercase}
  \defaultfontfeatures[\rmfamily]{Ligatures=TeX,Scale=1}
\fi
\IfFileExists{upquote.sty}{\usepackage{upquote}}{}
\IfFileExists{microtype.sty}{% use microtype if available
  \usepackage[]{microtype}
  \UseMicrotypeSet[protrusion]{basicmath} % disable protrusion for tt fonts
}{}
\makeatletter
\@ifundefined{KOMAClassName}{% if non-KOMA class
  \IfFileExists{parskip.sty}{%
    \usepackage{parskip}
  }{% else
    \setlength{\parindent}{0pt}
    \setlength{\parskip}{6pt plus 2pt minus 1pt}}
}{% if KOMA class
  \KOMAoptions{parskip=half}}
\makeatother
\usepackage{xcolor}
\IfFileExists{xurl.sty}{\usepackage{xurl}}{} % add URL line breaks if available
\IfFileExists{bookmark.sty}{\usepackage{bookmark}}{\usepackage{hyperref}}
\hypersetup{
  pdftitle={Using Machine Learning Techniques to Identify Key Risk Factors for Diabetes and Undiagnosed Diabetes},
  pdfauthor={Avraham Adler},
  pdflang={en-US},
  colorlinks=true,
  linkcolor=Maroon,
  filecolor=Maroon,
  citecolor=Blue,
  urlcolor=purple,
  pdfcreator={LaTeX via pandoc}}
\usepackage{graphicx}
\makeatletter
\def\maxwidth{\ifdim\Gin@nat@width>\linewidth\linewidth\else\Gin@nat@width\fi}
\def\maxheight{\ifdim\Gin@nat@height>\textheight\textheight\else\Gin@nat@height\fi}
\makeatother
\setkeys{Gin}{width=\maxwidth,height=\maxheight,keepaspectratio}
\makeatletter
\def\fps@figure{htbp}
\makeatother
\providecommand{\tightlist}{%
  \setlength{\itemsep}{0pt}\setlength{\parskip}{0pt}}
\makeatletter
\def\verbatim@nolig@list{}
\makeatother

\usepackage{amsmath}
\usepackage[dvipsnames]{xcolor}
\usepackage[singlespacing]{setspace}
\usepackage{float}
\usepackage{booktabs}
\usepackage{longtable}
\usepackage{array}
\usepackage{multirow}
\usepackage{wrapfig}
\usepackage{colortbl}
\usepackage{pdflscape}
\usepackage{tabu}
\usepackage{threeparttable}
\usepackage{threeparttablex}
\usepackage[normalem]{ulem}
\usepackage{makecell}
\usepackage{xcolor}
\ifxetex
  \usepackage{polyglossia}
  \setmainlanguage[variant=american]{english}
\else
  \usepackage[main=american]{babel}
\def\languageshorthands#1{}
\fi
\ifluatex
  \usepackage{selnolig}  % disable illegal ligatures
\fi
\newlength{\cslhangindent}
\newlength{\csllabelwidth}
\newenvironment{CSLReferences}[2] % #1 hanging-ident, #2 entry spacing
 {% don't indent paragraphs
  \setlength{\parindent}{0pt}
  \ifodd #1 \everypar{\setlength{\hangindent}{\cslhangindent}}\ignorespaces\fi
  \ifnum #2 > 0
  \setlength{\parskip}{#2\baselineskip}
  \fi
 }%
 {}
\usepackage{calc}

\newcommand{\CSLLeftMargin}[1]{\parbox[t]{\csllabelwidth}{#1}}
\newcommand{\CSLRightInline}[1]{\parbox[t]{\linewidth - \csllabelwidth}{#1}\break}

\title{Using Machine Learning Techniques to Identify Key Risk Factors
for Diabetes and Undiagnosed Diabetes}
\author{Avraham Adler}
\date{May 10, 2021}

\begin{document}
\maketitle
\begin{abstract}
This paper reviews a wide selection of machine learning models built to
predict both the presence of diabetes and the presence of undiagnosed
diabetes using eight years of National Health and Nutrition Examination
Survey (NHANES) data. Models are tuned and compared via their Brier
Scores. The most relevant variables of the best performing models are
then compared. A Support Vector Machine with a linear kernel performed
best for predicting diabetes, returning a Brier score of 0.0654 and an
AUROC of 0.9235 on the test set. An elastic net regression performed
best for predicting undiagnosed diabetes with a Brier score of 0.0294
and an AUROC of 0.9439 on the test set. Similar features appear
prominently in the models for both sets of models. Blood osmolality,
family history, the prevalance of various compounds, and hypertension
are key indicators for all diabetes risk. For undiagnosed diabetes in
particular, there are ethnicity or genetic components which arise as
strong correlates as well.

\end{abstract}

{
\hypersetup{linkcolor=}
\setcounter{tocdepth}{2}
\tableofcontents
}
\hypertarget{introduction}{%
\section{Introduction}\label{introduction}}

Diabetes is a chronic and lifelong disease with many health risks.
According to the American Diabetes Association (ADA)
\protect\hyperlink{ref-ADA2021C1}{{[}1{]}}, it:

\begin{quote}
\emph{\ldots poses a significant financial burden to individuals and
society. It is} \emph{estimated that the annual cost of diagnosed
diabetes in 2017 was \$327 billion,} \emph{including \$237 billion in
direct medical costs and \$90 billion in reduced} \emph{productivity.}
\end{quote}

There are a few types of diabetes of which the most common is type-2
diabetes \protect\hyperlink{ref-ADA2021W1}{{[}2{]}}. This is when the
human body develops a resistance to insulin over time leading to
elevated blood sugars \protect\hyperlink{ref-ADA2014Diagnosis}{{[}3{]}},
eventually causing heart, organ, and neural complications---if not
outright failure \protect\hyperlink{ref-ADA2021C10}{{[}4{]}},
\protect\hyperlink{ref-ADA2021C11}{{[}5{]}}. In 2010, a study published
in \emph{The Lancet} estimated that screening can reduce
diabetes-related complications, increase the number of quality-adjusted
life-years, and prevent a significant number of simulated deaths, with
early onset screening being more cost-effective than later
\protect\hyperlink{ref-Kahn.etal2010}{{[}6{]}}.

The actual diagnosis of diabetes is relatively straightforward using
either fasting plasma glucose (FPG) or hemoglobin A1C levels
\protect\hyperlink{ref-ADA2021C2}{{[}7{]}}. The FPG measures the blood
sugar level at a point in time. The A1C levels reflect the average sugar
in the blood over a three-month period
\protect\hyperlink{ref-CDC2018A1CW}{{[}8{]}}.

Diabetes diagnosis using machine learning is not a new problem; there
exists a body of research related to indicators of Type-2 diabetes. A
proprietary implementation of decision trees on over 22,000 records from
Iranian hospitals between 2009 and 2011 was developed by
\protect\hyperlink{ref-Habibi.etal2015}{{[}9{]}}.
\protect\hyperlink{ref-Lai.etal2019}{{[}10{]}} built a gradient-boosted
machine model on a proprietary dataset of 200,000 records of 13,000
Canadian patient data obtained from the Canadian Primary Care Sentinel
Surveillance Network (CPCSSN).
\protect\hyperlink{ref-MahboobAlam.etal2019}{{[}11{]}} used classic
machine learning techniques such as artificial neural networks, random
forests, and K-means clustering to develop a predictive model relating
various medical characteristics with diabetes using the formerly
public-access Pima Indians dataset.

This paper will use data from the National Health and Nutrition
Examination Survey (NHANES) to build a predictive model for diabetes.
What this study attempts beyond previous analyses of NHANES data, such
as \protect\hyperlink{ref-Yu.etal2010}{{[}12{]}} and
\protect\hyperlink{ref-Dinh.etal2019}{{[}13{]}}, is to build a second
model to specifically predict \emph{undiagnosed} type-2 diabetes and
compare the significant predictors between the two models. Logically,
correlates of undiagnosed diabetes may prove especially valuable for
early detection.

The remainder of this paper is organized as follows. First, the NHANES
dataset will be described. Following will be a description of feature
selection based on a concomitant literature review. Thereafter will be a
description of the general data preparation with sections depicting an
overview of each model. More detail about specific implementations may
be found as comments in the code appendix. Afterwards, the models will
be tested against the holdout set. The last section of the paper will be
a reprise of any findings and suggestions for further research. The body
of the paper will be followed by two appendices. The first will display
statistical tables and figures of secondary interest, and the second
will contain the R \protect\hyperlink{ref-R}{{[}14{]}} code used to
perform the analyses. The final element of the paper will be the
reference list.

\hypertarget{nhanes-data}{%
\section{NHANES Data}\label{nhanes-data}}

Among the many projects of The Center for Disease Control and Prevention
(CDC) National Center for Health Statistics (NCHS) is the National
Health and Nutrition Examination Survey (NHANES). This is:

\begin{quote}
\emph{a program of studies designed to assess the health and nutritional
status of} \emph{adults and children in the United States. The survey is
unique in that it} \emph{combines interviews and physical
examinations.\ldots The NHANES interview includes} \emph{demographic,
socioeconomic, dietary, and health-related questions. The}
\emph{examination component consists of medical, dental, and
physiological} \emph{measurements, as well as laboratory tests
administered by highly trained} \emph{medical personnel}
\protect\hyperlink{ref-HealthStatistics2017}{{[}15{]}}.
\end{quote}

The study comprises hundreds of questions and tests from each of
thousands of respondents. Most of the questions are grouped in
individual files and must be cross-referenced with both the general
demographic data as well as findings from other tests. For example, the
diabetes questionnaire is separate from the lab results for A1C or
glucose.

The benefit of using the NHANES data is that there are many features
which may be investigated for their relationship to diabetes. However,
this very flexibility has a drawback in the combinatorial nature of the
possibilities. A reasonable balance of records, fields, and constructed
features will be necessary. To include all available data would be both
impractical and intractable. Earlier studies
\protect\hyperlink{ref-Yu.etal2010}{{[}12{]}},
\protect\hyperlink{ref-Dinh.etal2019}{{[}13{]}},
\protect\hyperlink{ref-Dong.etal2011}{{[}16{]}} have identified
variables related to diabetes including the expected body mass index
(BMI), age, family history, and hypertension. This study will augment
these prior-identified variables with other judgmentally selected
variables to provide a wider universe to analyze.

The NHANES data has been collected in two-year cycles since 1999.
However, there have been a number of changes in both the data collected
and the variable names. For this study the data collected will be that
of the four cycles spanning 2011--2018. This is a reasonable balance
between volume of data and selection of features. While this prevents
some possible potentially valuable metrics---such as waist-to-hip ratio
(WHR) or iron metabolism---it allows for 39,156 records in the initial
dataset, as opposed to many fewer if limited to 2017--2018. After
cleaning and munging the data, however, the final data set has only
10,329 records which both fit the desired characteristics and have
sufficient features. Much of this reduction is due to the NHANES policy
of only performing laboratory work on a subset of all interviewed.
Reviewing prior literature and the data, 32 features were selected
against which to train the models. Some of these will be expanded into
dummy variables for models which require them.

\hypertarget{variable-description}{%
\section{Variable Description}\label{variable-description}}

\hypertarget{target-variables}{%
\subsection{Target Variables}\label{target-variables}}

There are two simultaneous modeling exercises which will be performed
for each algorithm---the first will train a model to detect the presence
of diabetes and the second will train a model for the presence of
\emph{undiagnosed} diabetes. Every model will be trained, validated, and
tested identically but on slightly different extracts of the data sets
as explained below.

\hypertarget{diabetes}{%
\subsubsection{Diabetes}\label{diabetes}}

The first target is defined similar to
\protect\hyperlink{ref-Semerdjian.Frank2017}{{[}17{]}}:

\begin{enumerate}
\def\labelenumi{\arabic{enumi}.}
\tightlist
\item
  If a patient answered ``yes'' to the question ``Have you ever been
  ever been told by a doctor or health professional that you have
  diabetes or sugar diabetes?''
\item
  If a patient's lab work exceeds the accepted thresholds for diabetes
  as per \protect\hyperlink{ref-ADA2021C2}{{[}7{]}}:

  \begin{enumerate}
  \def\labelenumii{\arabic{enumii}.}
  \tightlist
  \item
    Fasting plasma glucose (FPG) level of 126 mg/dL or greater
  \item
    Glycohemoglobin (A1C) level of 6.5\% or greater
  \end{enumerate}
\end{enumerate}

Using the results from the diabetes survey questions and lab work, each
observation will be tagged as either ``diabetic'' or ``non-diabetic.''
Among the 10,329 records, there are 8,245 patients without diabetes and
2,084 with diabetes.

\hypertarget{undiagnosed-diabetes}{%
\subsubsection{Undiagnosed Diabetes}\label{undiagnosed-diabetes}}

The second target is defined for cases where the blood sugar or A1C
threshold was breached but the answer to part 1. above was not ``Yes.''
This includes cases of ``I don't remember,'' ``Borderline,'' or ``No.''
The object of this modeling exercise is to identify predictors or
correlates for diabetics who have not as yet been diagnosed. The desire
is to identify if there are any correlates which stand out from those
already diagnosed and thus more likely to have sought treatment.
Therefore, the training set for predicting undiagnosed diabetes will
remove all observations with \emph{diagnosed} diabetes. The hypothesis
is leaving these observations in would train for what causes someone to
get diagnosed and not for correlates for diabetes prior to any medical
or dietary interventions. Therefore, the 1,486 patients with diagnosed
diabetes will be removed from the data, leaving 598 positives against
the 8,245 patients without diabetes at all. As the two data sets are no
longer identical, model comparisons may only be made within each target
and not between them.

\hypertarget{prediabetes}{%
\subsubsection{PreDiabetes}\label{prediabetes}}

The existing literature is conflicted when it comes to classification of
prediabetes. These are people whose FPG level falls on the range between
100 mg/DL and 125 mg/DL, or whose A1C either falls between 5.7\% and
6.4\% or exhibited a 10\% increase. The scheme in
\protect\hyperlink{ref-Semerdjian.Frank2017}{{[}17{]}} only focuses on
diabetes. In \protect\hyperlink{ref-Yu.etal2010}{{[}12{]}}, two schemes
are used: one where prediabetes is grouped with diabetes and one where
it is not. A similar technique was used by
\protect\hyperlink{ref-Dinh.etal2019}{{[}13{]}} where the first scheme
trained for true diabetics and the second stage removed those considered
positive for the first stage and trained for prediabetes.\footnote{There
  appears to be a contradiction in
  \protect\hyperlink{ref-Dinh.etal2019}{{[}13{]}} in that their first
  stage considers \textbf{either} a ``yes'' answer \textbf{or} a FPG of
  126 mg/dL or greater to be diagnosed diabetes. They remove the
  diagnosed diabetes from the dataset, but say that those who answer
  ``no'' but have a FPG of 126+ mg/dL are undiagnosed and are maintained
  in the dataset. As these are people with actual diabetes and not
  prediabetes, consistency implies they too should have been removed
  from the data for the second stage.} For the purposes of this study,
prediabetes will not be grouped with diabetes.

\hypertarget{class-imbalance}{%
\subsubsection{Class Imbalance}\label{class-imbalance}}

Both modeling exercises suffer from class imbalance. Where the target is
all patients with diabetes, the negative-to-positive ratio is
approximately 4:1. Where the target is undiagnosed diabetes, the
negative-to-positive ratio is approximately 14:1. There are many ways to
address class imbalance. These include artificial sampling methods,
applying case weights, and implementing a cost function among many
others. The interested reader is directed to chapter 16 of
\protect\hyperlink{ref-Kuhn.Johnson2018}{{[}18{]}} and to
\protect\hyperlink{ref-Menardi.Torelli2014}{{[}19{]}} for a more
complete approach. In this analysis, the effects of class imbalance will
be ameliorated by the use of a strictly proper scoring metric, which
will be discussed in the appropriate section.

\hypertarget{gender}{%
\subsection{Gender}\label{gender}}

As the focus of this investigation is Type-2 diabetes, the data will
exclude pregnant women and people below the age of 20 to remove possible
confusions with Type-1 or gestational diabetes
\protect\hyperlink{ref-Yu.etal2010}{{[}12{]}}. Women whose pregnancy
status is missing will be considered not pregnant.

\hypertarget{blood-sugar}{%
\subsection{Blood Sugar}\label{blood-sugar}}

Counter-intuitively, the values for FPG and A1C will not be included in
the training data. These are not predictors of diabetes; rather, they
define diabetes. This approach is consistent with
\protect\hyperlink{ref-Yu.etal2010}{{[}12{]}} and
\protect\hyperlink{ref-Dinh.etal2019}{{[}13{]}}, but is not consistent
with \protect\hyperlink{ref-Lai.etal2019}{{[}10{]}} or
\protect\hyperlink{ref-Dong.etal2011}{{[}16{]}} who use glycemic values
in their models.

\hypertarget{demographics}{%
\subsection{Demographics}\label{demographics}}

Race and ethnicity have been shown to be related to diabetes
\protect\hyperlink{ref-Dinh.etal2019}{{[}13{]}}. This stands to reason
for a disease with genetic components. Similarly, family history, where
known, is a valuable indicator. This study will use the enhanced racial
and ethnic profile implemented by NHANES in 2011 which separated Asians
from non-Hispanic whites
\protect\hyperlink{ref-NHANES_DEMO2013}{{[}20{]}}.

\hypertarget{income}{%
\subsection{Income}\label{income}}

The NHANES data measures income via categorical data. Moreover, there is
overlap between some of the categories. While categories--even
ordered--can be analyzed using modern techniques and software, it is
easier to convert the data to numerical values. This also allows for
easier imputation of the missing values. As the original data was
converted into ranges to begin with, it will be reconverted to numeric
using the midpoint of each range. For the categories representing the
simple cutoff at \$20,000 the lower category will use \$15,000 and the
higher category will use \$60,000---the rounded median of the median US
incomes over the eight year experience period
\protect\hyperlink{ref-Semega.etal2020}{{[}21{]}}.

\hypertarget{body-measurements}{%
\subsection{Body measurements}\label{body-measurements}}

A suite of measurements will be included in the initial feature set. It
is expected that BMI will remain one of the significant factors. While
the waist-to-hip ratio (WHR) is also a known measurement of interest,
NHANES only started capturing hip circumference in the 2017--2018 cycle.

\hypertarget{hypertension}{%
\subsection{Hypertension}\label{hypertension}}

The presence of hypertension is defined as per the 2017 revision by the
American College of Cardiology and the American Heart Association
\protect\hyperlink{ref-Whelton.etal2018}{{[}22{]}}. This lowered the
thresholds for hypertension to 130 mmHg for the systolic pressure and 80
mmHg for the diastolic pressure. Also, anyone actively taking blood
pressure medication is considered to be be hypertensive even if their
readings are below the thresholds---similar to the definition of
diabetes. When there is more than one measurement in the data, the
measurements will be averaged and that average used for identification
\protect\hyperlink{ref-Whelton.etal2018}{{[}22, Sec. 4.1{]}}.

\hypertarget{hyperlipidemia}{%
\subsection{Hyperlipidemia}\label{hyperlipidemia}}

Unlike with diabetes or hypertension, the targets for hyperlipidemia are
not fixed but depend on the presence of other positive of negative risk
factors \protect\hyperlink{ref-NCEP-ATIII_Summary2001}{{[}23{]}}, Table
4. Therefore, instead of a logical indicator for hyperlipidemia,
individual values for high-density lipoproteins (HDL)---the ``good''
cholesterol, low-density lipoproteins (LDL), and triglycerides will be
analyzed, although \protect\hyperlink{ref-Habibi.etal2015}{{[}9{]}} did
not consider triglycerides.

\hypertarget{smoking}{%
\subsection{Smoking}\label{smoking}}

Some studies have shown a relationship between smoking and an increased
prevalence for type-2 diabetes
\protect\hyperlink{ref-Dong.etal2011}{{[}16{]}}. The serum level of
cotinine will be used as a proxy for smoking and second-hand smoke
exposure, as cotinine has the longer half-life of the two primary
metabolites of nicotine \protect\hyperlink{ref-Benowitz1999}{{[}24{]}}.

\hypertarget{blood-measurements}{%
\subsection{Blood Measurements}\label{blood-measurements}}

As diabetes is measured by the level of sugar in the blood, many of the
variables from the NHANES standard blood assay, such as blood sodium,
potassium, and calcium, will be investigated for their relationship with
diabetes.

\hypertarget{model-preparation}{%
\section{Model Preparation}\label{model-preparation}}

\hypertarget{training-and-testing}{%
\subsection{Training and Testing}\label{training-and-testing}}

Approximately 25\% of the data for each modeling exercise---all diabetes
and undiagnosed diabetes---will be set aside as a testing set against
which the ``best-tuned'' models will be compared. This will be done
prior to any missing data imputation to prevent leakage of the test data
into the training data. The split will attempt to maintain the same
ratio of diabetics to non-diabetics in both training and testing sets.

\hypertarget{missing-data-imputation}{%
\subsection{Missing Data Imputation}\label{missing-data-imputation}}

Missing numeric variables will be imputed using a bagged tree method.
This is where a tree-based model is built for each predictor based on
the other predictors in the training set. This is a more powerful than a
k-nearest-neighbors approach and also does not require centering and
scaling of the data. This training-data based imputation model will be
used to impute missing data in the test sets.

\hypertarget{categorical-variables}{%
\subsection{Categorical Variables}\label{categorical-variables}}

Some of the models do not handle categorical variables well. For these
models, dummy variables will be created. Other models, specifically
tree-based models, may do better when categorical variables are not
converted to binary, as their splitting performs more efficiently when
treated as a single variable
\protect\hyperlink{ref-Kuhn.Johnson2018}{{[}18{]}}, Ch 14,
\protect\hyperlink{ref-Kuhn.Johnson2019}{{[}25, Sec. 57{]}}. The
training and testing sets will have both formats. As these both
represent the same underlying data, models which use either as input may
be properly compared one with the other.

\hypertarget{tuning-and-testing-metrics}{%
\subsection{Tuning and Testing
Metrics}\label{tuning-and-testing-metrics}}

The metric used to tune the hyperparameters---which will also be the
primary metric used to compare across the models---is the \textbf{Brier
score} \protect\hyperlink{ref-Brier1950}{{[}26{]}}. The Brier score is a
strictly proper scoring metric
\protect\hyperlink{ref-Selten1998}{{[}27{]}},
\protect\hyperlink{ref-CV90705}{{[}28{]}}, which means that it finds its
optimal value \emph{solely} at the true probabilities
\protect\hyperlink{ref-Merkle.Steyvers2013}{{[}29{]}}. One of the
benefits of using a strictly proper scoring metric for evaluation is
that it is much less affected by imbalanced data
\protect\hyperlink{ref-HarrellJr.2015}{{[}30{]}} as it is a
probabilistic measure and not a threshold measure
\protect\hyperlink{ref-Brownlee2020}{{[}31{]}}. For the binary
classification problem, the Brier score simplifies to the mean squared
error between the predicted probability and the actual event coded as 0
if it did not occur or 1 if it did
\protect\hyperlink{ref-Harrell2020}{{[}32{]}}.

Other metrics which will be calculated and displayed include the area
under the ROC (AUROC), the area under the precision-recall curve (AUPRC)
using the interpolation method of
\protect\hyperlink{ref-Davis.Goadrich2006}{{[}33{]}}, accuracy, balanced
accuracy, precision, recall, F score, and the Matthews correlation
coefficient. Most of these metrics are well-known in machine learning
but two deserve more explanation in light of the unbalanced nature of
the classes.

\emph{Balanced accuracy} is the arithmetic mean of sensitivity (recall)
and specificity. Thus, it is the mean of the ratio of true positives to
all positives and true negatives to all negatives. In cases where the
decision boundary puts all predictions on the majority side there will
either be no false positives or no false negatives, making one ratio
will be 1 and the other 0. The balanced accuracy in this case will be
50\%, notwithstanding the prevalence of the majority class
\protect\hyperlink{ref-Brodersen.etal2010}{{[}34{]}}.

The \emph{Matthews correlation coefficient} (MCC) is a single metric
which combines all four quadrants of the confusion matrix into a single
score ranging between -1 and 1
\protect\hyperlink{ref-Chicco.etal2021}{{[}35{]}}. First introduced by
Matthews \protect\hyperlink{ref-Matthews1975}{{[}36{]}}, it can also be
viewed as a correlation measure between the the observations and the
predictions. Mathematically:

\[
\textrm{MCC} = \frac{TP \cdot TN - FP\cdot FN}{\sqrt{(TP+FP)\cdot(TP+FN)
\cdot(TN+FP)\cdot(TN+FN)}}
\]

where \(TP, TN, FP,\) and \(FN\) are the true positive, true negative,
false positive, and false negative counts respectively. In cases where
the selection is completely random, such as all the observations being
predicted as the same class, the convention is to set the denominator to
1 which lets the fraction equal 0 as well.

\hypertarget{hyperparameter-tuning}{%
\subsection{Hyperparameter Tuning}\label{hyperparameter-tuning}}

A five times repeated ten-fold cross-validation (CV) method will be used
to tune the hyperparameters for each model. This is where the average of
five different ten-fold CVs are used to select the tuning parameters.
This is not the same as fifty-fold CV although fifty models are run.
This allows for the reduced bias of ten fold CV and a further reduced
variance around the metrics due to the repeats. As the Brier metric is
optimized through minimization, reducing the process variance around
metrics of order of 0.05 should help better identify optimal parameter
sets. The \texttt{caret} package for R
\protect\hyperlink{ref-caret}{{[}37{]}} will be used to train and test
models for which it has interfaces. Otherwise, models will use their
native packages.

\hypertarget{feature-selection}{%
\subsection{Feature Selection}\label{feature-selection}}

All models will start with the full feature set and most will undergo
feature selection. Some models, such as the elastic net, have built-in
feature selection. Those which do not may undergo a reverse recursive
feature elimination where the model starts saturated and then
algorithmically removes features. Some models do not have the
implementation to undergo feature elimination; this will be noted in the
model description sections.

\hypertarget{predictive-models}{%
\section{Predictive Models}\label{predictive-models}}

The models which will be investigated are:

\begin{itemize}
\tightlist
\item
  Regression

  \begin{itemize}
  \tightlist
  \item
    Logistic regression with reverse recursive feature elimination
  \item
    Logistic regression with step-wise AIC feature elimination
  \item
    Elastic Net regression
  \end{itemize}
\item
  Linear Discriminant Analysis (LDA)

  \begin{itemize}
  \tightlist
  \item
    LDA with reverse recursive feature elimination
  \end{itemize}
\item
  Support Vector Machines (SVM)

  \begin{itemize}
  \tightlist
  \item
    SVM with a linear kernel on all features
  \end{itemize}
\item
  Naïve Bayes (NB)

  \begin{itemize}
  \tightlist
  \item
    NB with with reverse recursive feature elimination
  \end{itemize}
\item
  Neural Networks

  \begin{itemize}
  \tightlist
  \item
    Multilayer perceptron using Keras and Tensorflow
  \end{itemize}
\item
  Decision Trees

  \begin{itemize}
  \tightlist
  \item
    CART
  \item
    C5.0
  \end{itemize}
\item
  Ensemble Methods

  \begin{itemize}
  \tightlist
  \item
    Random Forest using the \texttt{ranger} package
  \item
    Extreme Gradient Boosting using XGBoost
  \item
    Random Forest using LightGBM
  \item
    Extreme Gradient Boosting using LightGBM
  \end{itemize}
\end{itemize}

\hypertarget{regression}{%
\subsection{Regression}\label{regression}}

Logistic regression will be used with three kinds of feature reduction
techniques. The first, and most general, is based on a simple wrapper
loop which starts at the fully saturated model and removes predictors
based on their (lack of) importance
\protect\hyperlink{ref-caret}{{[}37{]}}. This is a technique useful for
many model families.

The second feature selection technique used will be based on the model's
Akaike information criterion (AIC). This is a likelihood-based metric
which reflects how close the model is to a theoretical ``best'' model,
which may or may not be in the set of models reviewed
\protect\hyperlink{ref-Burnham.Anderson2002}{{[}38{]}} Ch. 6.4. The
model starts fully saturated and then algorithmically removes or adds
predictors based on how they reduce the AIC of the model, as for AIC,
the lower the score, the better.

The third technique is to use an linear elastic net model. This is a
blend between ridge and lasso regression. The lasso portion of the net
can truly eliminate variables and the ridge portion of the net can drive
variables close together or asymptotically close to zero
\protect\hyperlink{ref-Friedman.etal2010}{{[}39{]}},
\protect\hyperlink{ref-Zou.Hastie2005}{{[}40{]}}.

\hypertarget{linear-discrminant-analysis}{%
\subsection{Linear Discrminant
Analysis}\label{linear-discrminant-analysis}}

Linear discriminant analysis (LDA) will be tried with recursive feature
elimination. In the binary case, LDA works under the assumption that
there is a linear boundary between two normally-distributed random
variables \protect\hyperlink{ref-ESL}{{[}41{]}} Ch. 4.3.

\hypertarget{support-vector-machines-svm}{%
\subsection{Support Vector Machines
(SVM)}\label{support-vector-machines-svm}}

Support Vector Machines (SVM) theoretically operate by ``boosting'' the
problem into a high-enough dimensionality that a separating hyperplane
may be found. This plane, when projected back into the dimensionality of
the feature space, may be highly non-linear. In practice, however, the
kernel trick is used to calculate the needed metrics without having to
calculate any of the n-dimensional distances
\protect\hyperlink{ref-ESL}{{[}41{]}} Ch. 12.3.

Unfortunately, the recursive feature elimination routines are not
implemented for the SVM in the \texttt{caret} framework for R. While
there has been research in implementing such processes for SVM in
biogenetics \protect\hyperlink{ref-Sanz.etal2018}{{[}42{]}}, the model
here will be trained on all the features. The linear kernel outperformed
the radial basis function kernel in all attempts on this data set so the
latter will not be shown.

\hypertarget{nauxefve-bayes}{%
\subsection{Naïve Bayes}\label{nauxefve-bayes}}

Generally, naïve Bayes (NB) does better than expected in classification
problems but the expectation here is muted. The uncorrelated assumptions
may be too big of a jump for this dataset; moreover, NB tends to perform
poorly with unbalanced data due to a bias which reduces the weights for
underrepresented classes
\protect\hyperlink{ref-Rennie.etal2003}{{[}43{]}}. The recursive
elimination wrapper in the \texttt{caret} package will be used here for
feature elimination. The implementation for NB uses the area under the
ROC curve to decide variable importance.

\hypertarget{neural-networks}{%
\subsection{Neural Networks}\label{neural-networks}}

Neural networks are well-established predictive modeling methods which,
through today's computing power, have evolved into deep learning
techniques. This analysis leverages the Keras
\protect\hyperlink{ref-Chollet2015}{{[}44{]}} and Tensorflow
\protect\hyperlink{ref-Abadi.etal2015}{{[}45{]}} deep learning
implementations. More advanced neural network techniques, such as
convolutional or recurrent neural networks, are geared towards data with
multiple dimensions such as the temporal or spatial. Here, the data is
treated as singularly dimensional, so dense neural network connections
will be used. Also, instead of implementing a specialized Brier score
for the Keras/Tensorflow framework, the log-loss function---which is
also a strictly proper scoring metric---will be used for tuning. The
best models will still be compared via their Brier score performance on
the testing set.

\hypertarget{decision-trees}{%
\subsection{Decision Trees}\label{decision-trees}}

The first decision tree will be a classification and regression tree
(CART). One of the benefits of a single CART is that it can be plotted.
It is unlikely to be the best model, but it will demonstrate the
built-in feature selection by having fewer than 32 nodes.

\includegraphics{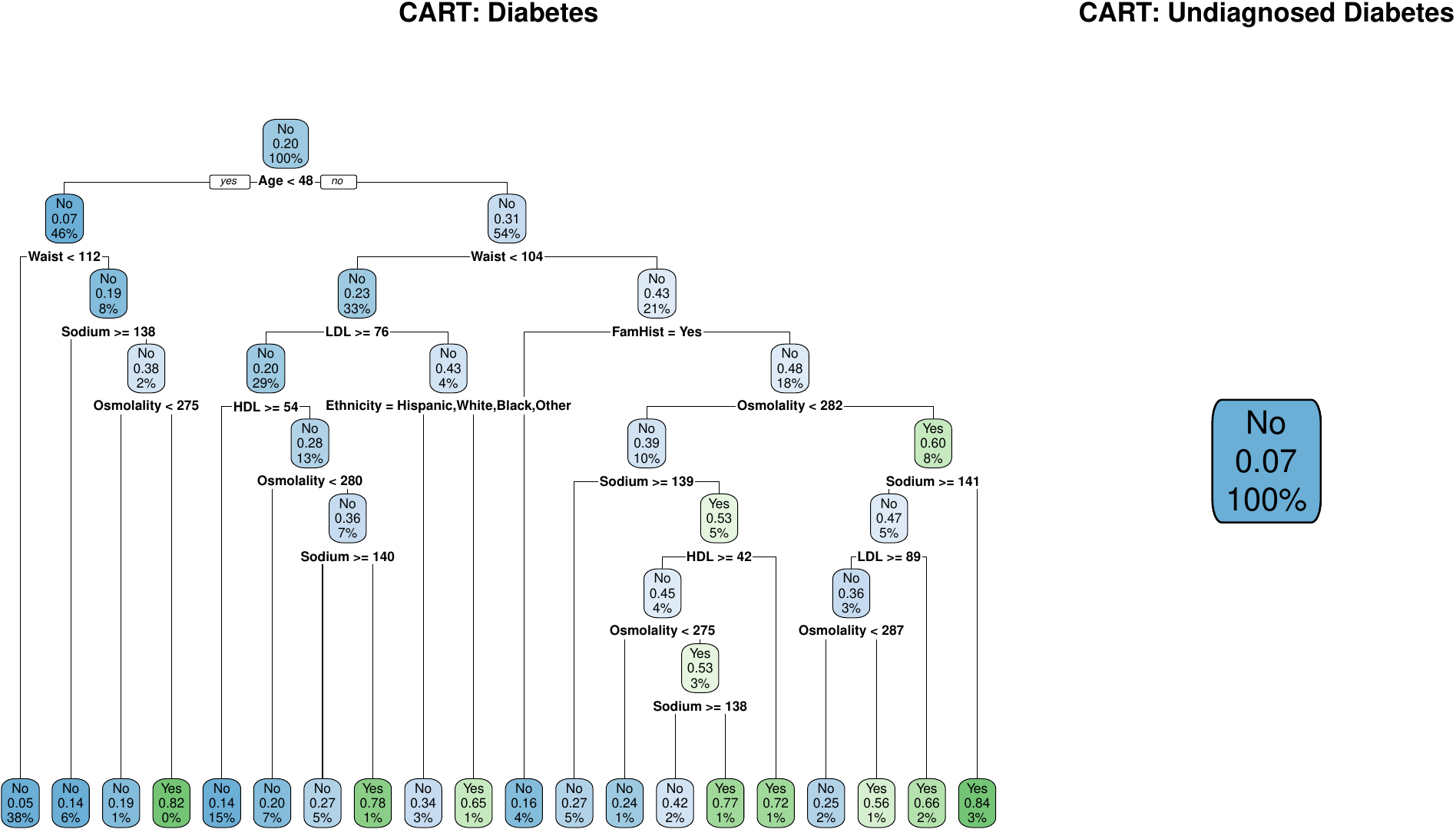}

As seen in the dendrograms above, The CART for predicting diabetes
focuses on age, waist size, family history, blood sodium levels, LDL
cholesterol, and blood osmolality as the more important variables. The
CART for undiagnosed diabetes is overwhelmed by the imbalance and simply
assigns everything to the larger class.

A more sophisticated classification tree-building technique is that of
C5.0, an upgrade to Quinlan's C4.5
\protect\hyperlink{ref-Quinlan2014}{{[}46{]}}. Here, the method has the
ability to winnow the features. It also can enhance the tree structure
by allowing a rules-based model to be the node and not just a value. The
best fitting C5.0 trees did not prune any features and used a
rules-based method, so there is no dendrogram for either the diabetes
model or the undiagnosed diabetes model.

\hypertarget{random-forest}{%
\subsection{Random Forest}\label{random-forest}}

Random forest is a decision tree based method which uses an ensemble of
trees to reduce the propensity to overfit. It is similar to bagging in
that the final results are based on a collection of bootstrapped trees
fit to a random sampling of the data. It improves on bagging by using a
random set of features to decide on each split
\protect\hyperlink{ref-ESL}{{[}41{]}} Ch 15,
\protect\hyperlink{ref-Breiman2001}{{[}47{]}}. This section uses the
random forest implementation found in the \texttt{ranger}
\protect\hyperlink{ref-ranger}{{[}48{]}} package for R. It also
implements the feature elimination routine of
\protect\hyperlink{ref-Deng.Runger2012}{{[}49{]}}. This adds a
regularization penalty to the splitting routine with the intention of
preventing deeper trees which overfit the training data.

\hypertarget{extreme-gradient-boosting}{%
\subsection{Extreme Gradient Boosting}\label{extreme-gradient-boosting}}

Boosted trees are another decision tree based ensemble method. Unlike
bagged trees and random forests, boosted trees work by repeatedly
training a tree against the errors from a prior iteration
\protect\hyperlink{ref-ESL}{{[}41{]}} Ch. 10. The extreme gradient
boosting algorithm (XGB) is a particularly efficient implementation of
boosted trees \protect\hyperlink{ref-Chen.Guestrin2016}{{[}50{]}}. Two
types of booster---the weak learners used in each iteration training
against the errors---will be used: a linear-type learner akin to a
penalized regression and a tree-type learner akin to a CART.

\hypertarget{lightgbm}{%
\subsection{LightGBM}\label{lightgbm}}

The LightGBM software package is an implementation of boosted decision
trees with a number of design differences from the XGB implementation.
For example, it grows the tree leaf-wise (vertically) over level-wise
(horizontally). There are other design choices which make it especially
suitable for large data \protect\hyperlink{ref-LightGBM2017}{{[}51{]}}.
A random forest ensemble, a boosted tree, and a piecewise linear
gradient boosted tree will be built for each target. This last is
similar to the LightGBM boosted tree and XGBTree, but instead of each
node being a constant, each node is itself a linear model
\protect\hyperlink{ref-Shi.etal2019}{{[}52{]}}.

\hypertarget{results}{%
\section{Results}\label{results}}

To compare the results of a model's ability to predict versus its
ability to learn, and to observe if there is severe overfitting, it is
customary to show the model's training statistics and not only its
testing statistics. These statistics may be found in the statistical
appendix. \clearpage

\hypertarget{predicting-diabetes}{%
\subsection{Predicting Diabetes}\label{predicting-diabetes}}

\hypertarget{performance}{%
\subsubsection{Performance}\label{performance}}

\begin{table}[H]

\caption{\label{tab:tableD}Results for Predicting Diabetes}
\centering
\resizebox{\linewidth}{!}{
\begin{tabular}[t]{lrrrrrrrrr}
\toprule
Model & Brier & AUPRC & AUROC & Acc & BalAcc & MCC & Precision & Recall & F\\
\midrule
SVM & 0.06540 & 0.85685 & 0.92347 & 0.91441 & 0.82233 & 0.71798 & 0.87879 & 0.66795 & 0.75900\\
Logistic Red. & 0.06596 & 0.85770 & 0.93005 & 0.91208 & 0.82876 & 0.71229 & 0.84670 & 0.68906 & 0.75979\\
Logistic Step & 0.06651 & 0.85574 & 0.93076 & 0.91441 & 0.83524 & 0.72090 & 0.84722 & 0.70250 & 0.76810\\
Elastic Net & 0.06652 & 0.85598 & 0.93079 & 0.91441 & 0.83452 & 0.72070 & 0.84884 & 0.70058 & 0.76761\\
LGBM: Bst LinTree & 0.07094 & 0.84365 & 0.93069 & 0.90511 & 0.80718 & 0.68567 & 0.85025 & 0.64299 & 0.73224\\
\addlinespace
LDA & 0.07436 & 0.84387 & 0.93050 & 0.89969 & 0.76578 & 0.66381 & 0.93377 & 0.54127 & 0.68530\\
XGB: Tree & 0.07536 & 0.82283 & 0.92700 & 0.89620 & 0.79372 & 0.65493 & 0.82025 & 0.62188 & 0.70742\\
LGBM: Boost Tree & 0.07832 & 0.81295 & 0.91952 & 0.89311 & 0.80253 & 0.64997 & 0.78291 & 0.65067 & 0.71069\\
XGB: Linear & 0.08034 & 0.80360 & 0.92116 & 0.88962 & 0.78816 & 0.63410 & 0.78922 & 0.61804 & 0.69322\\
C5.0 & 0.09144 & 0.78577 & 0.91470 & 0.87994 & 0.77492 & 0.60199 & 0.75545 & 0.59885 & 0.66809\\
\addlinespace
Rand F. Red. & 0.09728 & 0.73473 & 0.89473 & 0.86987 & 0.71913 & 0.54809 & 0.80731 & 0.46641 & 0.59124\\
LGBM: RandF & 0.11348 & 0.64706 & 0.86470 & 0.84508 & 0.66345 & 0.44130 & 0.73913 & 0.35893 & 0.48320\\
CART & 0.12011 & 0.57015 & 0.80614 & 0.84314 & 0.67658 & 0.44358 & 0.69463 & 0.39731 & 0.50549\\
Naïve Bayes Red. & 0.13238 & 0.47178 & 0.79966 & 0.80713 & 0.61243 & 0.29253 & 0.54182 & 0.28599 & 0.37437\\
K/TF DenseNN & 0.20364 & 0.20313 & 0.50084 & 0.79822 & 0.50000 & 0.00000 & NA & 0.00000 & NA\\
\bottomrule
\end{tabular}}
\end{table}

The table above compares the performance of the 15 trained models. All
models, with the exception of the dense neural network, improved on the
``no-information rate'' (NIR) accuracy of 0.7982. The top four models
performed very similarly in all metrics and outperformed the others in
almost all metrics. It is interesting that the top performing models are
all some form of linear model. The three logistic models are linear in
sigmoid space and the SVM is linear in a hyperspace. Logistic models
being superior to other models in the cases of class imbalance was also
noted by Menardi \& Torelli
\protect\hyperlink{ref-Menardi.Torelli2014}{{[}19, p. 22{]}}. Linear
discriminant analysis did not perform as well. While it is linear in the
sense that it searches for a linear boundary, it assumes that the
variables being separated have Gaussian behavior which does seem to be
the case here. The next best set of models are the various flavors of
gradient boosted trees, random forests, and trailing them the
non-ensemble based trees. The naïve Bayes classifier provides barely any
gain over selecting the majority class 100\% of the time and the dense
neural network does not learn anything at all.

\hypertarget{important-variables}{%
\subsubsection{Important Variables}\label{important-variables}}

As the best performing model is a SVM, there is no actual model formula
which can be displayed. In general, predictive models outside the realm
of regressions often do not have direct formulaic representation. One
measure used to compare variables in predictive models is called
variable importance (VI). While the mathematics of calculating this
variable differs for every model family, it allows for comparison across
models---albeit imperfect
\protect\hyperlink{ref-Groemping2009}{{[}53{]}}. Focusing on the four
best-performing models helps reduce complexities. The VI of the three
linear regression models is based on the absolute value of the
t-statistic for the predictors, and for the SVM model it is based on the
AUROC.

\begin{table}[H]

\caption{\label{tab:tableImpD}Diabetes Prediction: Variable Importance}
\centering
\begin{tabular}[t]{llll}
\toprule
SVM & Red. Logit & StepAIC Logit & ENet\\
\midrule
Age & Osmolality & Age & Sodium\\
Waist & Sodium & Waist & Osmolality\\
Osmolality & UreaNitrogen & Osmolality & FamHist.Yes\\
Hypertension.Yes & FamHist.Yes & Hypertension.Yes & Ethnicity.White\\
BMI & Age & BMI & UreaNitrogen\\
\addlinespace
Triglycerides & Ethnicity.White & Triglycerides & Hypertension.Yes\\
UreaNitrogen & Waist & UreaNitrogen & Albumin\\
HDL & Hypertension.Yes & HDL & Phosphorus\\
Weight & LDL & Weight & Calcium\\
ArmC & LegLen & ArmC & Gender.Female\\
\addlinespace
Albumin & Phosphorus & Albumin & Ethnicity.Hispanic\\
Alk.Phosphate & Asp.Aminotransferase & Alk.Phosphate & Bilirubin\\
\bottomrule
\end{tabular}
\end{table}

In the table above, the variables are listed in order of their
importance within each model. It is clear that the most important
features correlated with diabetes include age, blood osmolality, blood
sodium, waist size/BMI or both, family history, hypertension, and
cholesterol. The second-best model is the logistic regression that
underwent recursive feature reduction. Therefore, the important
variables and the direction of their influence can be seen directly from
the model formula.

\begin{table}[H]

\caption{\label{tab:bestRegFormD}Reduced Logit Model}
\centering
\begin{tabular}[t]{lrrrr}
\toprule
  & Estimate & Std. Error & z value & Pr(>|z|)\\
\midrule
(Intercept) & -20.000 & 2.616 & -7.647 & 0.000\\
Osmolality & 1.549 & 0.052 & 29.504 & 0.000\\
Sodium & -2.934 & 0.100 & -29.394 & 0.000\\
UreaNitrogen & -0.521 & 0.020 & -26.541 & 0.000\\
FamHist.Yes & -1.081 & 0.132 & -8.166 & 0.000\\
\addlinespace
Age & 0.027 & 0.003 & 8.371 & 0.000\\
Ethnicity.White & -0.836 & 0.094 & -8.890 & 0.000\\
Waist & 0.025 & 0.003 & 9.400 & 0.000\\
Hypertension.Yes & 0.527 & 0.101 & 5.198 & 0.000\\
LDL & -0.006 & 0.001 & -5.151 & 0.000\\
\addlinespace
LegLen & -0.042 & 0.012 & -3.599 & 0.000\\
Phosphorus & 0.275 & 0.080 & 3.455 & 0.001\\
\bottomrule
\end{tabular}
\end{table}

Increased blood osmolality and decreasing sodium and nitrogen levels
have the highest z-values. Dr.~Valentine Burroughs, Director of
Endocrinology, Diabetes and Metabolism for Montefiore Nyack Hospital,
explained that what is occurring is that the larger sugar molecules are
lodging in the blood, pushing out the smaller sodium and nitrogen
compounds, making the blood thicker as well
\protect\hyperlink{ref-Burroughs2021}{{[}54{]}}.

\hypertarget{comparison-with-prior-studies}{%
\subsubsection{Comparison with Prior
Studies}\label{comparison-with-prior-studies}}

The models reviewed here compare favorably with others built on NHANES
data, although it must be recognized that different years may have been
used for these studies. The study in
\protect\hyperlink{ref-Yu.etal2010}{{[}12{]}} used NHANES data from
1999--2004. Their best performing model for detecting any diabetes, also
a SVM, had an AUROC of 0.8347. As seen in Table 1, twelve of the models
in this study exceeded that score.

The study in \protect\hyperlink{ref-Dinh.etal2019}{{[}13{]}} did not use
a holdout set to test the effectiveness of their model. Rather, they
used downsampling to create a balanced sample from the data and split
that 80/20 to validate their models. They then used ten-fold
cross-validation of the entire dataset to evaluate the model's
performance, which returned an AUROC of 0.962. For comparison, the
average AUROC over ten-fold cross-validation of the entire dataset for
the best-performing SVM model was 0.921. This AUROC is based on the
original distribution of observations in the training set, of course.

\hypertarget{predicting-undiagnosed-diabetes}{%
\subsection{Predicting Undiagnosed
Diabetes}\label{predicting-undiagnosed-diabetes}}

\hypertarget{performance-1}{%
\subsubsection{Performance}\label{performance-1}}

\begin{table}[H]

\caption{\label{tab:tableU}Results for Predicting Undiagnosed Diabetes}
\centering
\resizebox{\linewidth}{!}{
\begin{tabular}[t]{lrrrrrrrrr}
\toprule
Model & Brier & AUPRC & AUROC & Acc & BalAcc & MCC & Precision & Recall & F\\
\midrule
Elastic Net & 0.02937 & 0.74705 & 0.94393 & 0.96471 & 0.80052 & 0.69002 & 0.81982 & 0.61074 & 0.70000\\
Logistic Step & 0.02941 & 0.74548 & 0.94332 & 0.96471 & 0.79429 & 0.68759 & 0.83178 & 0.59732 & 0.69531\\
Logistic Red. & 0.02993 & 0.73691 & 0.93573 & 0.96109 & 0.76745 & 0.64839 & 0.81818 & 0.54362 & 0.65323\\
SVM & 0.03027 & 0.73284 & 0.93808 & 0.96244 & 0.80864 & 0.67776 & 0.77049 & 0.63087 & 0.69373\\
LDA & 0.03465 & 0.72782 & 0.94107 & 0.95430 & 0.68909 & 0.55712 & 0.86364 & 0.38255 & 0.53023\\
\addlinespace
LGBM: Bst LinTree & 0.03508 & 0.67523 & 0.92087 & 0.95882 & 0.73510 & 0.61659 & 0.84524 & 0.47651 & 0.60944\\
XGB: Tree & 0.03784 & 0.63129 & 0.91049 & 0.95339 & 0.68861 & 0.54772 & 0.83824 & 0.38255 & 0.52535\\
LGBM: Boost Tree & 0.03789 & 0.63029 & 0.90994 & 0.95249 & 0.68812 & 0.53870 & 0.81429 & 0.38255 & 0.52055\\
XGB: Linear & 0.04184 & 0.57441 & 0.89965 & 0.94751 & 0.64498 & 0.46674 & 0.80000 & 0.29530 & 0.43137\\
C5.0 & 0.04922 & 0.52163 & 0.88758 & 0.94118 & 0.56376 & 0.34634 & 1.00000 & 0.12752 & 0.22619\\
\addlinespace
Rand F. Red. & 0.05013 & 0.43340 & 0.87269 & 0.93891 & 0.54698 & 0.29696 & 1.00000 & 0.09396 & 0.17178\\
LGBM: RandF & 0.05473 & 0.30638 & 0.84009 & 0.93303 & 0.52826 & 0.16222 & 0.52941 & 0.06040 & 0.10843\\
Naïve Bayes Red. & 0.05919 & 0.18254 & 0.76421 & 0.93258 & 0.50000 & 0.00000 & NA & 0.00000 & NA\\
CART & 0.06288 & 0.06742 & 0.50000 & 0.93258 & 0.50000 & 0.00000 & NA & 0.00000 & NA\\
K/TF DenseNN & 0.14567 & 0.07291 & 0.49612 & 0.93258 & 0.50000 & 0.00000 & NA & 0.00000 & NA\\
\bottomrule
\end{tabular}}
\end{table}

Developing a successful model for undiagnosed diabetes proved more
difficult. This is almost certainly due to the gross imbalance within
the classes. Not unexpectedly, the same group of four models proved
themselves superior to the remainder.

\hypertarget{important-variables-1}{%
\subsubsection{Important Variables}\label{important-variables-1}}

\begin{table}[H]

\caption{\label{tab:tableImpU}Undiagnosed Diabetes Prediction: Variable Importance}
\centering
\begin{tabular}[t]{llll}
\toprule
ENet & Red. Logit & StepAIC Logit & SVM\\
\midrule
Sodium & Osmolality & Waist & Waist\\
Osmolality & Sodium & Age & Age\\
UreaNitrogen & UreaNitrogen & Triglycerides & Triglycerides\\
Ethnicity.Asian & NA & BMI & BMI\\
Gender.Female & NA & Osmolality & Osmolality\\
\addlinespace
Hypertension.Yes & NA & Hypertension.Yes & Hypertension.Yes\\
FamHist.Yes & NA & HDL & HDL\\
Potassium & NA & ArmC & ArmC\\
Protein & NA & Weight & Weight\\
Calcium & NA & Alk.Phosphate & Alk.Phosphate\\
\addlinespace
Ethnicity.White & NA & UricAcid & UricAcid\\
Albumin & NA & Asp.Aminotransferase & Asp.Aminotransferase\\
\bottomrule
\end{tabular}
\end{table}

In the table above, the variables are listed in order of their
importance within each model. The variables are similar to those for the
all-diabetes model but there is less emphasis on body measurements and
more on gender and ethnic variables.

For undiagnosed diabetes, the best performing model is the elastic net
with \(\alpha =\) 1 and \(\lambda =\) 0. In other words, a pure ridge
regression. Therefore, there is no expectation that features will be
completely eliminated; however, their coefficients may be driven to 0.
While there is no simple formulaic representation of the elastic net,
there is a plotting method which shows how the variables behave as the
penalties change.

\includegraphics{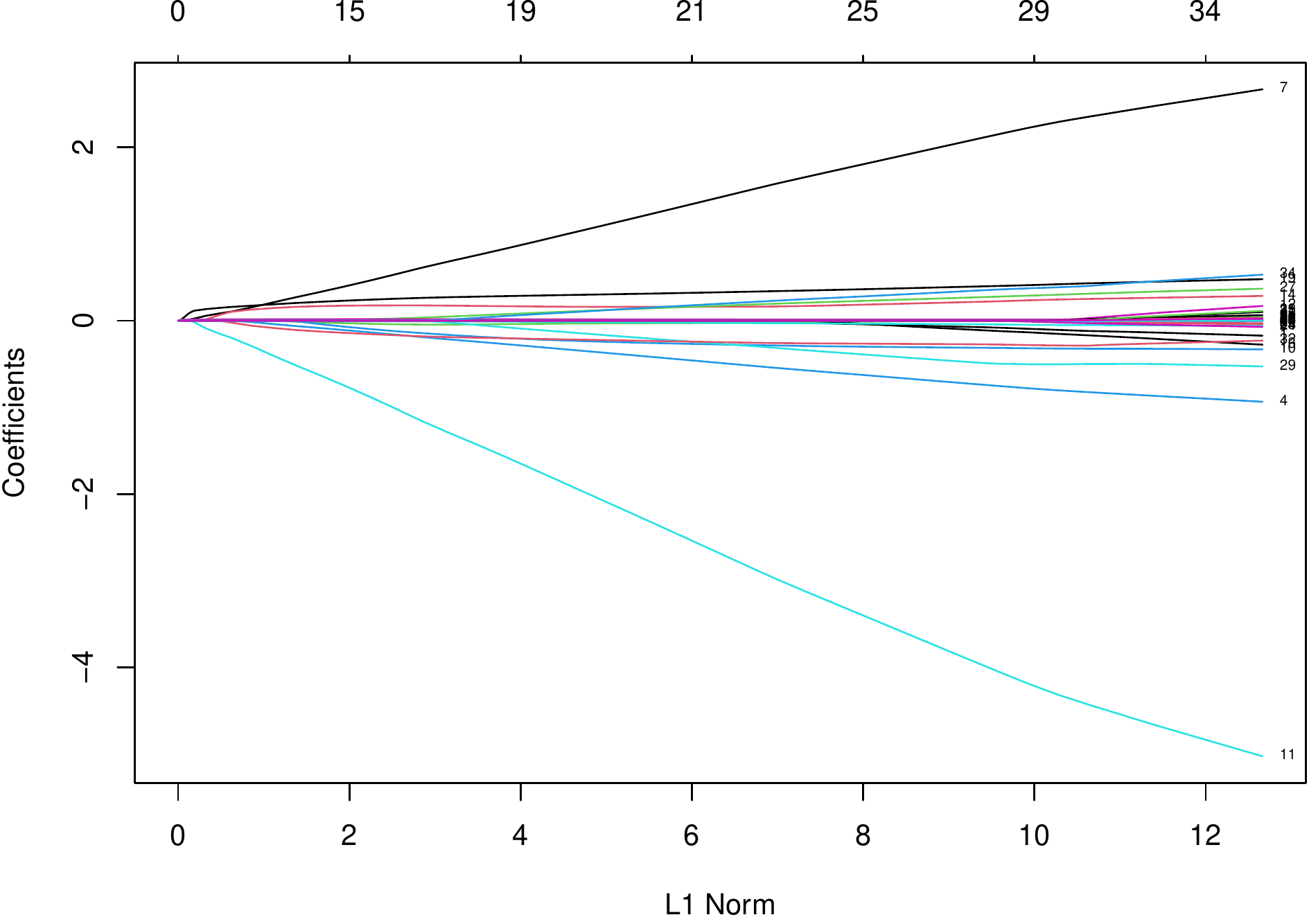}

Variables above the 0-line, such as 7, 34, and 19, are positively
correlated with an increased risk of undiagnosed diabetes. Variables
below 0, such as 11, 4, and 29 are inversely correlated with that risk.
The further to the left these variables ``break away'' from the
horizontal axis, the more important they are at that level of
regularization. As the graph traverses towards the right, more variables
are allowed in the model. Therefore, the importance of any variable may
change---or even switch direction---if other, possibly confounding,
variables are now explicitly modeled.

The best-performing model did not eliminate any of the variables---it is
the rightmost point in the above plot. However, we can still see that as
the model developed, there are no variables which switch direction but
there are those that change their slope and become more or less
``important.'' The twelve most important variables of the final model,
six positively correlated and six negatively correlated, are shown in
the table below.

\begin{table}[H]

\caption{\label{tab:enetUtbl}Key Variables in ENet for Undiagnosed Diabetes}
\centering
\begin{tabular}[t]{rrc}
\toprule
VarNum & Name & Correlation\\
\midrule
7 & Osmolality & +\\
34 & Ethnicity.Asian & +\\
19 & Hypertension.Yes & +\\
27 & FamHist.Yes & +\\
14 & Protein & +\\
\addlinespace
12 & Bilirubin & +\\
11 & Sodium & -\\
4 & UreaNitrogen & -\\
29 & Gender.Female & -\\
10 & Potassium & -\\
\addlinespace
13 & Calcium & -\\
32 & Ethnicity.White & -\\
\bottomrule
\end{tabular}
\end{table}

An interesting observation from the graph is the behavior of variable
19---hypertension. At first, it is a significant positive indicator of
undiagnosed diabetes. However, as more variables enter the model,
variable 34---Asian ethnicity---overtakes it in importance. With the
elastic net, the magnitude of \(\lambda\) determines the intensity of
the feature selection. The optimization procedure adjusts the lasso
penalty \(\lambda\) and the ridge penalty \(\alpha\) to find the best
model. Investigating the details of the optimization routine,
hypertension enters the model early on, when \(\lambda\) is relatively
large. It entered the model together with blood osmolality and only
three other variables preceded them in importance: waist size, age, and
triglycerides. Asian ethnicity enters the model after 15 variables, at
which point hypertension's coefficient is the second largest magnitude
in absolute value after osmolality. However, at the optimal model with
all variables included, hypertension's magnitude is eclipsed by Asian
ethnicity. Apparently, as other variables enter the model, they assume
some of the predictive burden of hypertension and it, in and of itself,
has less overall importance.

\hypertarget{conclusions-and-next-steps}{%
\section{Conclusions and Next Steps}\label{conclusions-and-next-steps}}

Using eight years of data from NHANES allows for the training of
accurate predictive models of type-2 diabetes. The family of linear
models performed better than those based on Gaussian assumptions or
trees, likely due to the imbalance of the data. The models in this study
compare very favorably to those of
\protect\hyperlink{ref-Yu.etal2010}{{[}12{]}} and are similar in
magnitude to those of \protect\hyperlink{ref-Dinh.etal2019}{{[}13{]}}
without having to change the empirical distribution of the data through
artificial resampling. The correlates most relevant for predicting
diabetes are age, weight, family history, various blood components,
cholesterol, and hypertension.

Focusing on the prediction of undiagnosed type-2 diabetes, the
correlates are similar, but give more weight to gender and ethnicity,
such as Asian being positively correlated and non-Hispanic whites and
females being negatively correlated. Remembering the dictum that
``correlation is not causation,'' and with the lessons of Simpson's
paradox in mind, it would be unwise to state that this is proof of a
biological component in women or white males which is absent in Asians.
The correlation may simply be who has better access to medical care, as
there may be a preponderance of white males diagnosed with diabetes and
thus absent from this data. However, the study does suggest that someone
with a family history of type-2 diabetes, active hypertension, high
blood osmolality, or low blood sodium would be wise to discuss
investigating their blood sugar and A1C levels with their physician.

A remaining disappointment is that the multilayer perceptron did not
learn anything in either modeling exercise. This may be due to the
imbalance or another structural issue with the data. Future research in
this topic should include testing the effects of artificial balancing on
some of these models, although that appears to be of less appeal to the
statistician than it is to the data scientist. Statisticians, such as
\protect\hyperlink{ref-Harrell2020a}{{[}55{]}} and
\protect\hyperlink{ref-Matloff2015}{{[}56{]}}, look at rebalancing the
data with a jaundiced eye. This is in contrast to the approach of more
the data science-aligned, such as
\protect\hyperlink{ref-Kuhn.Johnson2018}{{[}18{]}},
\protect\hyperlink{ref-Ramyachitra.Manikandan2014}{{[}57{]}}, and the
writings on popular data science portals such as
\texttt{towardsdatascience.com}
\protect\hyperlink{ref-Soni2018}{{[}58{]}} and \texttt{KDnuggets}
\protect\hyperlink{ref-Agarwal2020}{{[}59{]}}. These sources consider
rebalancing the data to be an acceptable, if not first step, in
addressing class imbalance.

Another area of future research would be to address the imbalance
through Bayesian methods. By creating a hierarchical Bayesian model with
reasonable priors on the probabilities of the positive and negative
classes, a more accurate model using the full joint distribution may be
obtained \protect\hyperlink{ref-Gelman.etal2014}{{[}60{]}} Ch. 16.
\clearpage

\hypertarget{statistical-appendix}{%
\section{Statistical Appendix}\label{statistical-appendix}}

\hypertarget{training-statistics}{%
\subsection{Training Statistics}\label{training-statistics}}

\begin{table}[H]

\caption{\label{tab:tableTrain}Training (Learning) Results for Predicting Diabetes}
\centering
\resizebox{\linewidth}{!}{
\begin{tabular}[t]{lrrrrrrrrr}
\toprule
Model & Brier & AUPRC & AUROC & Acc & BalAcc & MCC & Precision & Recall & F\\
\midrule
LGBM: Boost Tree & 0.04986 & 0.92195 & 0.96924 & 0.93443 & 0.86116 & 0.78724 & 0.92099 & 0.73832 & 0.81960\\
LGBM: Bst LinTree & 0.06267 & 0.87886 & 0.94825 & 0.91881 & 0.82436 & 0.73276 & 0.90679 & 0.66603 & 0.76798\\
Logistic Step & 0.06912 & 0.84601 & 0.92654 & 0.90861 & 0.81272 & 0.69778 & 0.86137 & 0.65195 & 0.74217\\
SVM & 0.06977 & 0.84140 & 0.91694 & 0.90987 & 0.79735 & 0.70074 & 0.91738 & 0.60871 & 0.73111\\
Logistic Red. & 0.07049 & 0.84033 & 0.92152 & 0.90874 & 0.81219 & 0.69835 & 0.86497 & 0.65032 & 0.74132\\
\addlinespace
Elastic Net & 0.07052 & 0.84124 & 0.92367 & 0.90799 & 0.81061 & 0.69563 & 0.86336 & 0.64735 & 0.73904\\
XGB: Tree & 0.07880 & 0.81057 & 0.91690 & 0.89550 & 0.78548 & 0.65036 & 0.83548 & 0.60103 & 0.69838\\
LDA & 0.08004 & 0.82589 & 0.92187 & 0.89343 & 0.75059 & 0.64001 & 0.92976 & 0.51110 & 0.65806\\
XGB: Linear & 0.08618 & 0.77756 & 0.90669 & 0.88414 & 0.76598 & 0.60979 & 0.80142 & 0.56789 & 0.66381\\
C5.0 & 0.09513 & 0.76284 & 0.90123 & 0.88106 & 0.76931 & 0.60241 & 0.77373 & 0.58196 & 0.66346\\
\addlinespace
LGBM: RandF & 0.10271 & 0.73592 & 0.89718 & 0.86498 & 0.69455 & 0.52465 & 0.83968 & 0.40883 & 0.54991\\
Rand F. Red. & 0.10275 & 0.69963 & 0.87850 & 0.86051 & 0.69585 & 0.50833 & 0.79212 & 0.41982 & 0.54742\\
CART & 0.13016 & 0.51410 & 0.77105 & 0.82982 & 0.64629 & 0.38299 & 0.65179 & 0.33859 & 0.44421\\
Naïve Bayes Red. & 0.13465 & 0.47006 & 0.78878 & 0.80870 & 0.60551 & 0.28736 & 0.55647 & 0.26486 & 0.35790\\
K/TF DenseNN & 0.20366 & 0.19687 & 0.49518 & 0.79824 & 0.50000 & 0.00000 & NA & 0.00000 & NA\\
\bottomrule
\end{tabular}}
\end{table}

\begin{table}[H]

\caption{\label{tab:tableTrain}Training (Learning) Results for Predicting Undiagnosed Diabetes}
\centering
\resizebox{\linewidth}{!}{
\begin{tabular}[t]{lrrrrrrrrr}
\toprule
Model & Brier & AUPRC & AUROC & Acc & BalAcc & MCC & Precision & Recall & F\\
\midrule
LGBM: Bst LinTree & 0.02283 & 0.87434 & 0.98174 & 0.96970 & 0.79166 & 0.73128 & 0.94604 & 0.58575 & 0.72352\\
LGBM: Boost Tree & 0.02482 & 0.86098 & 0.98144 & 0.96502 & 0.75714 & 0.68233 & 0.93927 & 0.51670 & 0.66667\\
Logistic Step & 0.02720 & 0.78165 & 0.95431 & 0.96593 & 0.79170 & 0.69758 & 0.86319 & 0.59020 & 0.70106\\
Logistic Red. & 0.02785 & 0.77153 & 0.94406 & 0.96563 & 0.78243 & 0.69179 & 0.88093 & 0.57056 & 0.68974\\
SVM & 0.02838 & 0.76767 & 0.94770 & 0.96520 & 0.80307 & 0.69610 & 0.82897 & 0.61556 & 0.70442\\
\addlinespace
Elastic Net & 0.02855 & 0.76623 & 0.94947 & 0.96451 & 0.78411 & 0.68419 & 0.85692 & 0.57546 & 0.68593\\
LDA & 0.03421 & 0.76204 & 0.94670 & 0.95607 & 0.69410 & 0.57774 & 0.90824 & 0.39110 & 0.54342\\
XGB: Tree & 0.03857 & 0.62558 & 0.91674 & 0.95176 & 0.68798 & 0.53326 & 0.80617 & 0.38291 & 0.51508\\
LGBM: RandF & 0.04074 & 0.66951 & 0.92136 & 0.94693 & 0.61628 & 0.45147 & 0.92920 & 0.23385 & 0.37367\\
XGB: Linear & 0.04707 & 0.48434 & 0.87795 & 0.94316 & 0.60368 & 0.39375 & 0.81018 & 0.21105 & 0.33233\\
\addlinespace
C5.0 & 0.05202 & 0.45552 & 0.85926 & 0.93879 & 0.55365 & 0.29352 & 0.90097 & 0.10820 & 0.19494\\
Rand F. Red. & 0.05310 & 0.37470 & 0.82704 & 0.93581 & 0.52934 & 0.21186 & 0.89826 & 0.05922 & 0.11433\\
Naïve Bayes Red. & 0.06102 & 0.13725 & 0.71806 & 0.93216 & 0.49992 & -0.00104 & 0.00000 & 0.00000 & NaN\\
CART & 0.06236 & 0.11744 & 0.56752 & 0.93216 & 0.51664 & 0.08201 & 0.48522 & 0.03606 & 0.12210\\
K/TF DenseNN & 0.14586 & 0.06831 & 0.50273 & 0.93231 & 0.50000 & 0.00000 & NA & 0.00000 & NA\\
\bottomrule
\end{tabular}}
\end{table}

What should be noted here is the behavior of the LightGBM tree models.
Their training errors are lower than the best-performing models on the
test set, and much lower than their own performance on the test set.
This is an example of overfitting. It may be that deliberately limiting
the training performance of these models to prevent overfitting may
return better results on the true holdout set.

\hypertarget{roc-and-pr}{%
\subsection{ROC and PR}\label{roc-and-pr}}

While neither the area under the ROC curve (AUROC) nor the area under
the PR curve (AUPRC) are proper scoring metrics, they are both widely
used for classification models. The AUROC in particular has a weaknesses
when the data is unbalanced. This weakness stems from the AUROC's
dependence on both the true positive and true negative rate. In cases
where the true positive rate is poor, the metric may still return good
values if dominated by the performance of the true negative class. The
AUPRC is less prone to this distortion as both precision and recall have
only true positive in their numerator. Therefore if the positive class
is much smaller, poor performance will not be not be overwhelmed by
excellent performance on the negative class
\protect\hyperlink{ref-Goadrich.etal2006}{{[}61{]}}. Nevertheless, it is
customary to return not only the single area under the curve values but
also the ROC and PR curve plots.

\hypertarget{predicting-diabetes-1}{%
\subsubsection{Predicting Diabetes}\label{predicting-diabetes-1}}

\includegraphics{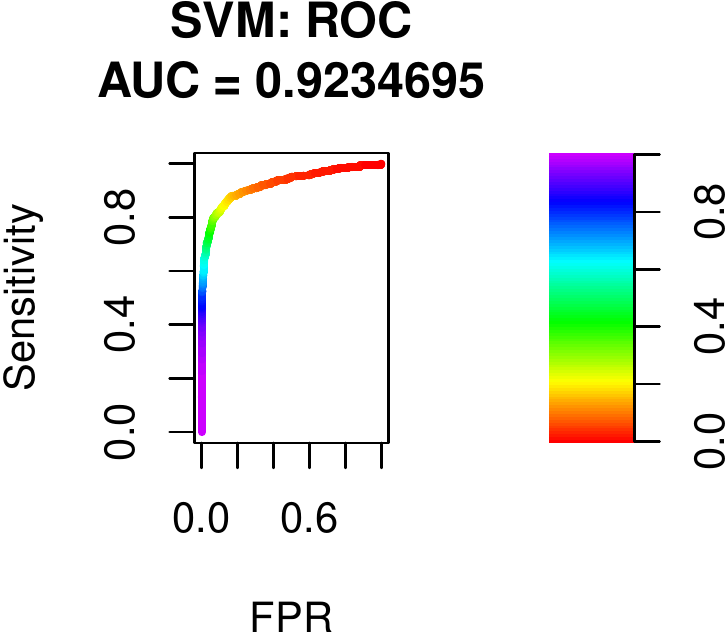}
\includegraphics{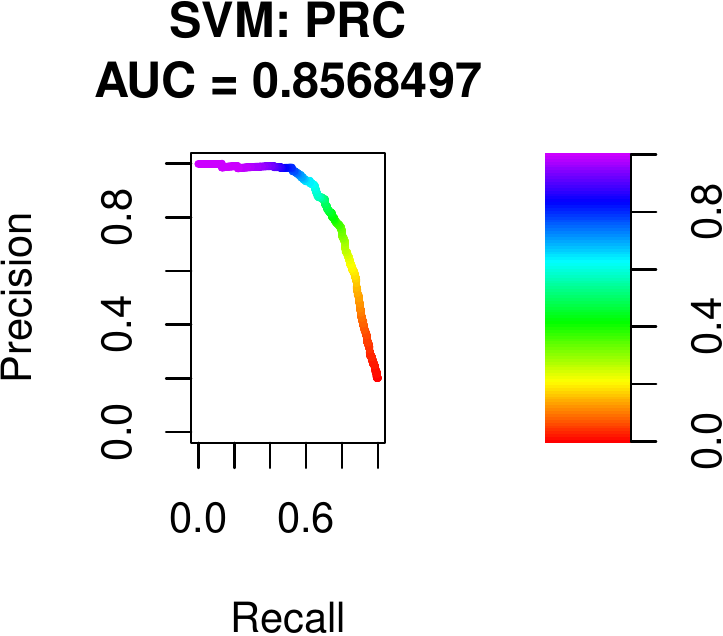}

\hypertarget{predicting-undiagnosed-diabetes-1}{%
\subsubsection{Predicting Undiagnosed
Diabetes}\label{predicting-undiagnosed-diabetes-1}}

\includegraphics{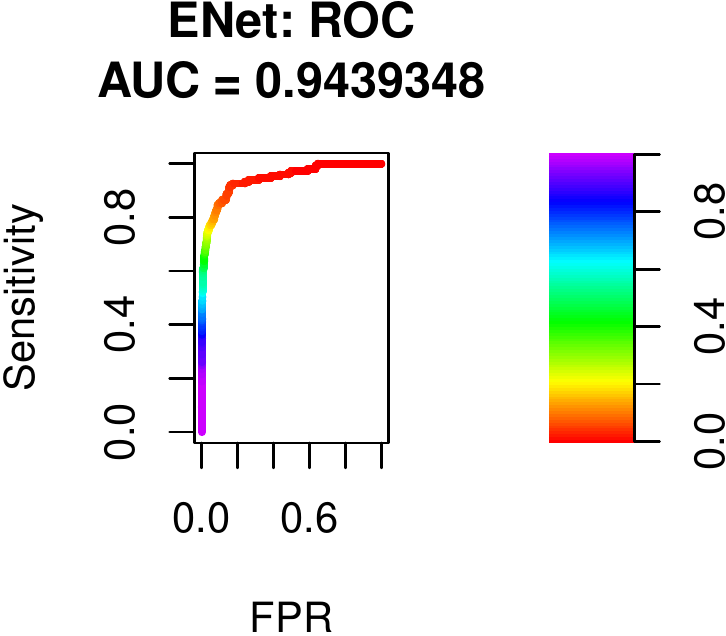}
\includegraphics{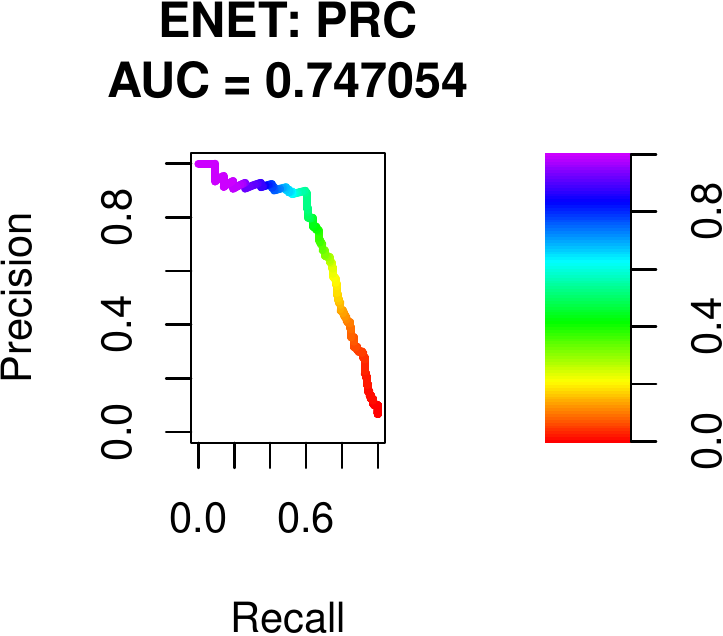}

\hypertarget{references}{%
\section*{References}\label{references}}
\addcontentsline{toc}{section}{References}

\hypertarget{refs}{}
\begin{CSLReferences}{0}{0}
\leavevmode\hypertarget{ref-ADA2021C1}{}%
\CSLLeftMargin{{[}1{]} }
\CSLRightInline{American Diabetes Association, {``Improving care and
promoting health in populations,''} \emph{Standards of Medical Care in
Diabetes---2021}, vol. 44, pp. S7--S14, 2021, doi:
\href{https://doi.org/10.2337/dc21-s001}{10.2337/dc21-s001}. }

\leavevmode\hypertarget{ref-ADA2021W1}{}%
\CSLLeftMargin{{[}2{]} }
\CSLRightInline{American Diabetes Association, {``The path to
understanding diabetes starts here,''} 2021. {[}Online{]}. Available:
\url{https://www.diabetes.org/diabetes}. {[}Accessed: 14-Mar-2021{]}}

\leavevmode\hypertarget{ref-ADA2014Diagnosis}{}%
\CSLLeftMargin{{[}3{]} }
\CSLRightInline{American Diabetes Association, {``Diagnosis and
classification of diabetes mellitus,''} \emph{Diabetes Care}, vol. 37,
no. Supplement 1, pp. S81--S90, 2014. }

\leavevmode\hypertarget{ref-ADA2021C10}{}%
\CSLLeftMargin{{[}4{]} }
\CSLRightInline{American Diabetes Association, {``Cardiovascular disease
and risk management,''} \emph{Standards of Medical Care in
Diabetes---2021}, vol. 44, pp. S125--S150, 2021, doi:
\href{https://doi.org/10.2337/dc21-s010}{10.2337/dc21-s010}. }

\leavevmode\hypertarget{ref-ADA2021C11}{}%
\CSLLeftMargin{{[}5{]} }
\CSLRightInline{American Diabetes Association, {``Microvascular
complications and foot care,''} \emph{Standards of Medical Care in
Diabetes---2021}, vol. 44, pp. S151--S167, 2021, doi:
\href{https://doi.org/10.2337/dc21-s011}{10.2337/dc21-s011}. }

\leavevmode\hypertarget{ref-Kahn.etal2010}{}%
\CSLLeftMargin{{[}6{]} }
\CSLRightInline{R. Kahn \emph{et al.}, {``Age at initiation and
frequency of screening to detect type 2 diabetes: A cost-effectiveness
analysis,''} \emph{The Lancet}, vol. 375, no. 9723, pp. 1365--1374,
2010, doi: \url{https://doi.org/10.1016/S0140-6736(09)62162-0}.
{[}Online{]}. Available:
\url{https://www.sciencedirect.com/science/article/pii/S0140673609621620}.
{[}Accessed: 07-Feb-2021{]}}

\leavevmode\hypertarget{ref-ADA2021C2}{}%
\CSLLeftMargin{{[}7{]} }
\CSLRightInline{American Diabetes Association, {``Classification and
diagnosis of diabetes,''} \emph{Standards of Medical Care in
Diabetes---2021}, vol. 44, pp. S15--S33, 2021, doi:
\href{https://doi.org/10.2337/dc21-s002}{10.2337/dc21-s002}. }

\leavevmode\hypertarget{ref-CDC2018A1CW}{}%
\CSLLeftMargin{{[}8{]} }
\CSLRightInline{Center for Disease Control and Prevention, {``All about
your A1C,''} 2018. {[}Online{]}. Available:
\url{https://www.cdc.gov/diabetes/managing/managing-blood-sugar/a1c.html}.
{[}Accessed: 14-Mar-2021{]}}

\leavevmode\hypertarget{ref-Habibi.etal2015}{}%
\CSLLeftMargin{{[}9{]} }
\CSLRightInline{S. Habibi, M. Ahmadi, and S. Alizadeh, {``Type 2
diabetes mellitus screening and risk factors using decision tree:
Results of data mining,''} \emph{Global Journal of Health Science}, vol.
7, no. 5, pp. 304--310, 2015, doi:
\href{https://doi.org/10.5539/gjhs.v7n5p304}{10.5539/gjhs.v7n5p304}.
{[}Online{]}. Available:
\url{https://www.ncbi.nlm.nih.gov/pmc/articles/PMC4803907/}.
{[}Accessed: 07-Feb-2021{]}}

\leavevmode\hypertarget{ref-Lai.etal2019}{}%
\CSLLeftMargin{{[}10{]} }
\CSLRightInline{H. Lai, H. Huang, K. Keshavjee, A. Guergachi, and X.
Gao, {``Predictive models for diabetes mellitus using machine learning
techniques,''} \emph{BMC Endocrine Disorders}, vol. 19, no. 101, 2019,
doi:
\href{https://doi.org/10.1186/s12902-019-0436-6}{10.1186/s12902-019-0436-6}.
}

\leavevmode\hypertarget{ref-MahboobAlam.etal2019}{}%
\CSLLeftMargin{{[}11{]} }
\CSLRightInline{T. Mahboob Alam \emph{et al.}, {``A model for early
prediction of diabetes,''} \emph{Informatics in Medicine Unlocked}, vol.
16, no. 100204, pp. 1--6, 2019, doi:
\url{https://doi.org/10.1016/j.imu.2019.100204}. {[}Online{]}.
Available:
\url{https://www.sciencedirect.com/science/article/pii/S2352914819300176}.
{[}Accessed: 07-Feb-2021{]}}

\leavevmode\hypertarget{ref-Yu.etal2010}{}%
\CSLLeftMargin{{[}12{]} }
\CSLRightInline{W. Yu, T. Liu, R. Valdez, M. Gwinn, and M. J. Khoury,
{``Application of support vector machine modeling for prediction of
common diseases: The case of diabetes and pre-diabetes,''} \emph{BMC
Medical Informatics and Decision Making}, vol. 10, pp. 1--7, 2010, doi:
\href{https://doi.org/10.1186/1472-6947-10-16}{10.1186/1472-6947-10-16}.
}

\leavevmode\hypertarget{ref-Dinh.etal2019}{}%
\CSLLeftMargin{{[}13{]} }
\CSLRightInline{A. Dinh, S. Miertschin, A. Young, and S. D. Mohantym,
{``A data-driven approach to predicting diabetes and cardiovascular
disease with machine learning,''} \emph{BMC Medical Informatics and
Decision Making}, vol. 19, no. 1, 1, pp. 211--225, 2019, doi:
\href{https://doi.org/10.1186/s12911-019-0918-5}{10.1186/s12911-019-0918-5}.
{[}Online{]}. Available:
\url{https://www.ncbi.nlm.nih.gov/pmc/articles/PMC6836338/}.
{[}Accessed: 14-Feb-2021{]}}

\leavevmode\hypertarget{ref-R}{}%
\CSLLeftMargin{{[}14{]} }
\CSLRightInline{R Core Team, \emph{R: A language and environment for
statistical computing}. Vienna, Austria: R Foundation for Statistical
Computing, 2021 {[}Online{]}. Available:
\url{http://www.R-project.org/}}

\leavevmode\hypertarget{ref-HealthStatistics2017}{}%
\CSLLeftMargin{{[}15{]} }
\CSLRightInline{National Center for Health Statistics, {``About the
national health and nutrition examination survey,''} 2017. {[}Online{]}.
Available: \url{https://www.cdc.gov/nchs/nhanes/about_nhanes.htm}.
{[}Accessed: 14-Feb-2021{]}}

\leavevmode\hypertarget{ref-Dong.etal2011}{}%
\CSLLeftMargin{{[}16{]} }
\CSLRightInline{J. Dong \emph{et al.}, {``Evaluation of a risk factor
scoring model in screening for undiagnosed diabetes in china
population,''} \emph{Journal of Zhejiang University SCIENCE B}, vol. 12,
no. 10, pp. 846--852, 2011, doi:
\href{https://doi.org/10.1631/jzus.B1000390}{10.1631/jzus.B1000390}. }

\leavevmode\hypertarget{ref-Semerdjian.Frank2017}{}%
\CSLLeftMargin{{[}17{]} }
\CSLRightInline{J. Semerdjian and S. Frank, {``An ensemble classifier
for predicting the onset ofType II diabetes,''} \emph{arXiv preprint},
pp. 1--8, 2017 {[}Online{]}. Available:
\url{https://arxiv.org/abs/1708.07480}. {[}Accessed: 14-Mar-2021{]}}

\leavevmode\hypertarget{ref-Kuhn.Johnson2018}{}%
\CSLLeftMargin{{[}18{]} }
\CSLRightInline{M. Kuhn and K. Johnson, \emph{Applied predictive
modeling}. Springer Science+Business Media, Inc., 2018. }

\leavevmode\hypertarget{ref-Menardi.Torelli2014}{}%
\CSLLeftMargin{{[}19{]} }
\CSLRightInline{G. Menardi and N. Torelli, {``Training and assessing
classification rules with imbalanced data,''} \emph{Data Mining and
Knowledge Discovery}, vol. 28, no. 1, pp. 92--122, 2014, doi:
\href{https://doi.org/10.1007/s10618-012-0295-5}{10.1007/s10618-012-0295-5}.
}

\leavevmode\hypertarget{ref-NHANES_DEMO2013}{}%
\CSLLeftMargin{{[}20{]} }
\CSLRightInline{N. Health and N. E. Survey, 2013. {[}Online{]}.
Available: \url{https://wwwn.cdc.gov/Nchs/Nhanes/2011-2012/DEMO_G.htm}.
{[}Accessed: 31-Mar-2021{]}}

\leavevmode\hypertarget{ref-Semega.etal2020}{}%
\CSLLeftMargin{{[}21{]} }
\CSLRightInline{J. Semega, M. Kollar, E. A. Shrider, and J. Creamer,
{``Income and poverty in the united states: 2019: Table a-2: Households
by total money income, race, and hispanic origin of householder: 1967 to
2019,''} U.S. Census Bureau, 2020 {[}Online{]}. Available:
\url{https://www2.census.gov/programs-surveys/demo/tables/p60/270/tableA2.xlsx}.
{[}Accessed: 01-Apr-2021{]}}

\leavevmode\hypertarget{ref-Whelton.etal2018}{}%
\CSLLeftMargin{{[}22{]} }
\CSLRightInline{P. K. Whelton \emph{et al.}, {``2017
ACC/AHA/AAPA/ABC/ACPM/AGS/APhA/ASH/ASPC/NMA/PCNA guideline for the
prevention, detection, evaluation, and management of high blood pressure
in adults,''} \emph{Journal of the American College of Cardiology}, vol.
71, no. 19, pp. e127--e248, 2017, doi:
\href{https://doi.org/10.1016/j.jacc.2017.11.006}{10.1016/j.jacc.2017.11.006}.
{[}Online{]}. Available:
\url{https://www.jacc.org/doi/abs/10.1016/j.jacc.2017.11.006}.
{[}Accessed: 31-Mar-2021{]}}

\leavevmode\hypertarget{ref-NCEP-ATIII_Summary2001}{}%
\CSLLeftMargin{{[}23{]} }
\CSLRightInline{Expert Panel on Detection Evaluation and T. of High
Blood Cholesterol in Adults, {``{Executive Summary of the Third Report
of the National Cholesterol Education Program (NCEP) Expert Panel on
Detection, Evaluation, and Treatment of High Blood Cholesterol in Adults
(Adult Treatment Panel III)},''} \emph{JAMA}, vol. 285, no. 19, pp.
2486--2497, 2001, doi:
\href{https://doi.org/10.1001/jama.285.19.2486}{10.1001/jama.285.19.2486}.
}

\leavevmode\hypertarget{ref-Benowitz1999}{}%
\CSLLeftMargin{{[}24{]} }
\CSLRightInline{N. L. Benowitz, {``Biomarkers of environmental tobacco
smoke exposure,''} \emph{Environmental Health Perspectives}, vol. 107,
pp. 349--355, 1999, doi:
\href{https://doi.org/10.1289/ehp.99107s2349}{10.1289/ehp.99107s2349}.
{[}Online{]}. Available:
\url{https://www.ncbi.nlm.nih.gov/pmc/articles/PMC1566286/}.
{[}Accessed: 31-Mar-2021{]}}

\leavevmode\hypertarget{ref-Kuhn.Johnson2019}{}%
\CSLLeftMargin{{[}25{]} }
\CSLRightInline{M. Kuhn and K. Johnson, \emph{Feature engineering and
selection: A practical approach for predictive models}. Taylor \&
Francis Ltd, 2020 {[}Online{]}. Available:
\url{https://bookdown.org/max/FES/}. {[}Accessed: 14-Apr-2021{]}}

\leavevmode\hypertarget{ref-Brier1950}{}%
\CSLLeftMargin{{[}26{]} }
\CSLRightInline{G. W. Brier, {``Verification of forecasts expressed in
terms of probability,''} \emph{Monthly weather review}, vol. 78, pp.
1--3, 1950 {[}Online{]}. Available:
\url{https://web.archive.org/web/20171023012737/https://docs.lib.noaa.gov/rescue/mwr/078/mwr-078-01-0001.pdf}.
{[}Accessed: 07-Feb-2021{]}}

\leavevmode\hypertarget{ref-Selten1998}{}%
\CSLLeftMargin{{[}27{]} }
\CSLRightInline{R. Selten, {``Axiomatic characterization of the
quadratic scoring rule,''} \emph{Experimental Economics}, vol. 1, no. 1,
pp. 43--61, 1998, doi:
\href{https://doi.org/10.1023/A:1009957816843}{10.1023/A:1009957816843}.
}

\leavevmode\hypertarget{ref-CV90705}{}%
\CSLLeftMargin{{[}28{]} }
\CSLRightInline{F. E. Harell, {``Why is AUC higher for a classifier that
is less accurate than for one that is more accurate?''} Cross Validated,
2014 {[}Online{]}. Available:
\url{https://stats.stackexchange.com/q/90705}. {[}Accessed:
07-Feb-2021{]}}

\leavevmode\hypertarget{ref-Merkle.Steyvers2013}{}%
\CSLLeftMargin{{[}29{]} }
\CSLRightInline{E. C. Merkle and M. Steyvers, {``Choosing a strictly
proper scoring rule,''} \emph{Decision Analysis}, vol. 10, no. 4, pp.
292--304, 2013, doi:
\href{https://doi.org/10.1287/deca.2013.0280}{10.1287/deca.2013.0280}.
{[}Online{]}. Available:
\url{https://www.researchgate.net/profile/Edgar_Merkle/publication/259715092_Choosing_a_Strictly_Proper_Scoring_Rule/links/00b7d52d70440e678e000000.pdf}.
{[}Accessed: 14-Apr-2021{]}}

\leavevmode\hypertarget{ref-HarrellJr.2015}{}%
\CSLLeftMargin{{[}30{]} }
\CSLRightInline{F. E. Harrell, \emph{Regression modeling strategies:
With applications to linear models, logistic and ordinal regression, and
survival analysis}, Second. Springer, 2015. }

\leavevmode\hypertarget{ref-Brownlee2020}{}%
\CSLLeftMargin{{[}31{]} }
\CSLRightInline{J. Brownlee, {``A gentle introduction to probability
metrics for imbalanced classification.''} 2020 {[}Online{]}. Available:
\url{https://machinelearningmastery.com/probability-metrics-for-imbalanced-classification/}.
{[}Accessed: 08-May-2021{]}}

\leavevmode\hypertarget{ref-Harrell2020}{}%
\CSLLeftMargin{{[}32{]} }
\CSLRightInline{F. E. Harrell, {``Damage caused by classification
accuracy and other discontinuous improper accuracy scoring rules.''}
Statistical Thinking, 2020 {[}Online{]}. Available:
\url{https://www.fharrell.com/post/class-damage/}. {[}Accessed:
15-Apr-2021{]}}

\leavevmode\hypertarget{ref-Davis.Goadrich2006}{}%
\CSLLeftMargin{{[}33{]} }
\CSLRightInline{J. Davis and M. Goadrich, {``The relationship between
precision-recall and ROC curves,''} in \emph{Proceedings of the 23rd
international conference on machine learning}, 2006, pp. 233--240, doi:
\href{https://doi.org/10.1145/1143844.1143874}{10.1145/1143844.1143874}
{[}Online{]}. Available:
\url{https://minds.wisconsin.edu/bitstream/handle/1793/60482/TR1551.pdf}}

\leavevmode\hypertarget{ref-Brodersen.etal2010}{}%
\CSLLeftMargin{{[}34{]} }
\CSLRightInline{K. H. Brodersen, C. S. Ong, K. E. Stephan, and J. M.
Buhmann, {``The balanced accuracy and its posterior distribution,''} in
\emph{20th international conference on pattern recognition}, 2010, pp.
3121--3124, doi:
\href{https://doi.org/10.1109/ICPR.2010.764}{10.1109/ICPR.2010.764}. }

\leavevmode\hypertarget{ref-Chicco.etal2021}{}%
\CSLLeftMargin{{[}35{]} }
\CSLRightInline{D. Chicco, N. Tötsch, and G. Jurman, {``The {Matthews}
correlation coefficient ({MCC}) is more reliable than balanced accuracy,
bookmaker informedness, and markedness in two-class confusion matrix
evaluation,''} \emph{{BioData} Mining}, vol. 14, no. 1, pp. 1--22, 2021,
doi:
\href{https://doi.org/10.1186/s13040-021-00244-z}{10.1186/s13040-021-00244-z}.
{[}Online{]}. Available:
\url{https://biodatamining.biomedcentral.com/articles/10.1186/s13040-021-00244-z}.
{[}Accessed: 25-Apr-2021{]}}

\leavevmode\hypertarget{ref-Matthews1975}{}%
\CSLLeftMargin{{[}36{]} }
\CSLRightInline{B. W. Matthews, {``Comparison of the predicted and
observed secondary structure of T4 phage lysozyme,''} \emph{Biochimica
et Biophysica Acta (BBA) - Protein Structure}, vol. 405, no. 2, pp.
442--451, 1975, doi: \url{https://doi.org/10.1016/0005-2795(75)90109-9}.
}

\leavevmode\hypertarget{ref-caret}{}%
\CSLLeftMargin{{[}37{]} }
\CSLRightInline{M. Kuhn, \emph{Caret: Classification and regression
training}. 2020 {[}Online{]}. Available:
\url{https://CRAN.R-project.org/package=caret}. {[}Accessed:
31-Mar-2021{]}}

\leavevmode\hypertarget{ref-Burnham.Anderson2002}{}%
\CSLLeftMargin{{[}38{]} }
\CSLRightInline{K. P. Burnham and D. R. Anderson, \emph{Model selection
and multimodel inference: A practical information-theoretic approach},
Second. New York: Springer Science+Business Media, Inc., 2002. }

\leavevmode\hypertarget{ref-Friedman.etal2010}{}%
\CSLLeftMargin{{[}39{]} }
\CSLRightInline{J. Friedman, T. Hastie, and R. Tibshirani,
{``Regularization paths for generalized linear models via coordinate
descent,''} \emph{Journal of Statistical Software}, vol. 33, no. 1, pp.
1--22, 2010 {[}Online{]}. Available:
\url{https://www.jstatsoft.org/v33/i01/}}

\leavevmode\hypertarget{ref-Zou.Hastie2005}{}%
\CSLLeftMargin{{[}40{]} }
\CSLRightInline{H. Zou and T. Hastie, {``Regularization and variable
selection via the elastic net,''} \emph{Journal of the Royal Statistical
Society: Series B (Statistical Methodology)}, vol. 67, no. 2, pp.
301--320, 2005, doi:
\url{https://doi.org/10.1111/j.1467-9868.2005.00503.x}. {[}Online{]}.
Available: \url{https://www.jstor.org/stable/pdf/3647580.pdf}.
{[}Accessed: 27-Apr-2021{]}}

\leavevmode\hypertarget{ref-ESL}{}%
\CSLLeftMargin{{[}41{]} }
\CSLRightInline{T. Hastie, R. Tibshirani, and J. Friedman, \emph{The
elements of statistical learning: Data mining, inference, and
prediction}, Second Ed. SPRINGER NATURE, 2009. }

\leavevmode\hypertarget{ref-Sanz.etal2018}{}%
\CSLLeftMargin{{[}42{]} }
\CSLRightInline{H. Sanz, C. Valim, E. Vegas, J. M. Oller, and F.
Reverter, {``{SVM}-{RFE}: Selection and visualization of the most
relevant features through non-linear kernels,''} \emph{BMC
Bioinformatics}, vol. 19, no. 432, pp. 1--18, 2018, doi:
\href{https://doi.org/10.1186/s12859-018-2451-4}{10.1186/s12859-018-2451-4}.
}

\leavevmode\hypertarget{ref-Rennie.etal2003}{}%
\CSLLeftMargin{{[}43{]} }
\CSLRightInline{J. D. Rennie, L. Shih, J. Teevan, and D. R. Karger,
{``Tackling the poor assumptions of naive bayes text classifiers,''} in
\emph{Proceedings of the 20th international conference on machine
learning (ICML-03)}, 2003, pp. 616--623. }

\leavevmode\hypertarget{ref-Chollet2015}{}%
\CSLLeftMargin{{[}44{]} }
\CSLRightInline{F. Chollet and others, {``Keras.''} https://keras.io,
2015 {[}Online{]}. Available: \url{https://keras.io}}

\leavevmode\hypertarget{ref-Abadi.etal2015}{}%
\CSLLeftMargin{{[}45{]} }
\CSLRightInline{M. Abadi \emph{et al.}, {``{TensorFlow}: Large-scale
machine learning on heterogeneous systems.''} 2015 {[}Online{]}.
Available: \url{https://www.tensorflow.org/}}

\leavevmode\hypertarget{ref-Quinlan2014}{}%
\CSLLeftMargin{{[}46{]} }
\CSLRightInline{J. R. Quinlan, \emph{C4.5: Programs for machine
learning}. San Mateo, Calif: Elsevier, 2014. }

\leavevmode\hypertarget{ref-Breiman2001}{}%
\CSLLeftMargin{{[}47{]} }
\CSLRightInline{L. Breiman, {``Random forests,''} \emph{Machine
Learning}, vol. 45, no. 1, pp. 5--32, 2001, doi:
\href{https://doi.org/10.1023/A:1010933404324}{10.1023/A:1010933404324}.
}

\leavevmode\hypertarget{ref-ranger}{}%
\CSLLeftMargin{{[}48{]} }
\CSLRightInline{M. N. Wright and A. Ziegler, {``{ranger}: A fast
implementation of random forests for high dimensional data in {C++} and
{R},''} \emph{Journal of Statistical Software}, vol. 77, no. 1, pp.
1--17, 2017, doi:
\href{https://doi.org/10.18637/jss.v077.i01}{10.18637/jss.v077.i01}. }

\leavevmode\hypertarget{ref-Deng.Runger2012}{}%
\CSLLeftMargin{{[}49{]} }
\CSLRightInline{H. Deng and G. Runger, {``Feature selection via
regularized trees,''} in \emph{The 2012 international joint conference
on neural networks ({IJCNN})}, 2012, pp. 1--8, doi:
\href{https://doi.org/10.1109/IJCNN.2012.6252640}{10.1109/IJCNN.2012.6252640}.
}

\leavevmode\hypertarget{ref-Chen.Guestrin2016}{}%
\CSLLeftMargin{{[}50{]} }
\CSLRightInline{T. Chen and C. Guestrin, {``{XGBoost}: A scalable tree
boosting system,''} in \emph{Proceedings of the 22nd {ACM} {SIGKDD}
international conference on knowledge discovery and data mining}, 2016,
pp. 785--794, doi:
\href{https://doi.org/10.1145/2939672.2939785}{10.1145/2939672.2939785}.
}

\leavevmode\hypertarget{ref-LightGBM2017}{}%
\CSLLeftMargin{{[}51{]} }
\CSLRightInline{G. Ke \emph{et al.}, {``LightGBM: A highly efficient
gradient boosting decision tree,''} \emph{Advances in Neural Information
Processing Systems}, vol. 30, pp. 3146--3154, 2017. }

\leavevmode\hypertarget{ref-Shi.etal2019}{}%
\CSLLeftMargin{{[}52{]} }
\CSLRightInline{Y. Shi, J. Li, and Z. Li, {``Gradient boosting with
piece-wise linear regression trees.''} 2019 {[}Online{]}. Available:
\url{https://arxiv.org/abs/1802.05640}. {[}Accessed: 27-Apr-2021{]}}

\leavevmode\hypertarget{ref-Groemping2009}{}%
\CSLLeftMargin{{[}53{]} }
\CSLRightInline{U. Grömping, {``Variable importance assessment in
regression: Linear regression versus random forest,''} \emph{The
American Statistician}, vol. 63, no. 4, pp. 308--319, 2009, doi:
\href{https://doi.org/10.1198/tast.2009.08199}{10.1198/tast.2009.08199}.
{[}Online{]}. Available: \url{https://doi.org/10.1198/tast.2009.08199}}

\leavevmode\hypertarget{ref-Burroughs2021}{}%
\CSLLeftMargin{{[}54{]} }
\CSLRightInline{V. J. Burroughs, personal communication, April 5, 2021.
}

\leavevmode\hypertarget{ref-Harrell2020a}{}%
\CSLLeftMargin{{[}55{]} }
\CSLRightInline{F. E. Harrell, {``Classification vs. prediction.''}
Statistical Thinking, 2020 {[}Online{]}. Available:
\url{https://www.fharrell.com/post/classification/}. {[}Accessed:
05-May-2021{]}}

\leavevmode\hypertarget{ref-Matloff2015}{}%
\CSLLeftMargin{{[}56{]} }
\CSLRightInline{N. Matloff, {``Unbalanced data is a problem? No,
BALANCED data is worse.''} Mad (Data) Scientist, 2015 {[}Online{]}.
Available:
\url{https://matloff.wordpress.com/2015/09/29/unbalanced-data-is-a-problem-no-balanced-data-is-worse/}.
{[}Accessed: 05-May-2021{]}}

\leavevmode\hypertarget{ref-Ramyachitra.Manikandan2014}{}%
\CSLLeftMargin{{[}57{]} }
\CSLRightInline{D. Ramyachitra and P. Manikandan, {``Imbalanced dataset
classification and solutions: A review,''} \emph{International Journal
of Computing and Business Research}, vol. 5, no. 4, pp. 1--29, 2014. }

\leavevmode\hypertarget{ref-Soni2018}{}%
\CSLLeftMargin{{[}58{]} }
\CSLRightInline{D. Soni, {``Dealing with imbalanced classes in machine
learning.''} towardsdatascience, 2018 {[}Online{]}. Available:
\url{https://towardsdatascience.com/dealing-with-imbalanced-classes-in-machine-learning-d43d6fa19d2}.
{[}Accessed: 05-May-2021{]}}

\leavevmode\hypertarget{ref-Agarwal2020}{}%
\CSLLeftMargin{{[}59{]} }
\CSLRightInline{R. Agarwal, {``The 5 most useful techniques to handle
imbalanced datasets.''} KDNuggets, 2020 {[}Online{]}. Available:
\url{https://www.kdnuggets.com/2020/01/5-most-useful-techniques-handle-imbalanced-datasets.html}.
{[}Accessed: 05-May-2021{]}}

\leavevmode\hypertarget{ref-Gelman.etal2014}{}%
\CSLLeftMargin{{[}60{]} }
\CSLRightInline{A. Gelman, J. B. Carlin, H. S. Stern, D. B. Dunson, A.
Vehtari, and D. B. Rubin, \emph{Bayesian data analysis, third edition}.
Chapman; Hall/CRC, 2014. }

\leavevmode\hypertarget{ref-Goadrich.etal2006}{}%
\CSLLeftMargin{{[}61{]} }
\CSLRightInline{M. Goadrich, L. Oliphant, and J. Shavlik, {``Gleaner:
Creating ensembles of first-order clauses to improve recall-precision
curves,''} \emph{Machine Learning}, vol. 64, no. 1, pp. 231--261, 2006,
doi:
\href{https://doi.org/10.1007/s10994-006-8958-3}{10.1007/s10994-006-8958-3}.
}

\end{CSLReferences}

\end{document}